\documentclass{article}
\usepackage{ijcai19}

\usepackage{times}
\usepackage{soul}
\usepackage{url}
\usepackage[hidelinks]{hyperref}
\usepackage[utf8]{inputenc}
\usepackage[small]{caption}
\usepackage{graphicx}
\usepackage{amsmath}
\usepackage{booktabs}
\usepackage{algorithm}
\usepackage{algorithmic}
\usepackage{subfigure}
\usepackage{multirow}
\usepackage{verbatim}
\begin{document}
\title{Improving the Robustness of Deep Neural Networks via\\ Adversarial Training with Triplet Loss}
\author{
Pengcheng Li$^1$\and
Jinfeng Yi$^2$\and
Bowen Zhou$^{2}$\And
Lijun Zhang$^1$\\
\affiliations
$^1$National Key Laboratory for Novel Software Technology Nanjing University, Nanjing 210023, China\\
$^2$JD AI Research, China\\
\emails
\{lipc, zhanglj\}@lamda.nju.edu.cn,
\{yijinfeng, bowen.zhou\}@jd.com
}
\maketitle
\begin{abstract}
Recent studies have highlighted that deep neural networks (DNNs) are vulnerable to adversarial examples. In this paper, we improve the robustness of DNNs by utilizing techniques of Distance Metric Learning. Specifically, we incorporate Triplet Loss, one of the most popular Distance Metric Learning methods, into the framework of adversarial training. Our proposed algorithm, Adversarial Training with Triplet Loss (AT$^2$L), substitutes the adversarial example against the current model for the anchor of triplet loss to effectively smooth the classification boundary. Furthermore, we propose an ensemble version of AT$^2$L, which aggregates different attack methods and model structures for better defense effects. Our empirical studies verify that the proposed approach can significantly improve the robustness of DNNs without sacrificing accuracy. Finally, we demonstrate that our specially designed triplet loss can also be used as a regularization term to enhance other defense methods.
    %and the robustness of the model with this regularization is better than that of the model after traditional adversarial training.  %Hence, this regularization can be applied to most of current defense methods for further improvement of robustness. 
    %We also extract our special designed triplet loss and take it as a regularization and the robustness of the model with this regularization can be no worse than that of the model after traditional adversarial training.
    %  The triple loss can also be used as a regularization term, and incorporated into other defense methods. The robustness of the model with this regularization shows better performance than that of the model after traditional adversarial training.
%\end{quote}
\end{abstract}

\section{Introduction}
Deep neural networks (DNNs) have been widely used for security-critical tasks, including but not limited to autonomous driving~\cite{DBLP:journals/corr/EvtimovEFKLPRS17}, surveillance~\cite{ouyang2013joint}, biometric recognition ~\cite{xu2017can}, and malware detection~\cite{yuan2014droid}. However, recent studies have shown that DNNs are vulnerable to adversarial examples~\cite{goodfellow2014explaining,papernot2016limitations,chen2017ead,ICDM:2018:Li}, which are carefully crafted instances that can mislead well-trained DNNs. This raises serious concerns about the security of machine learning models in many real-world applications.
%This raises a number of security concerns in real-world machine learning based applications. %Due to the wide varieties of methods currently used to attack machine learning models~\cite{huang2011adversarial,biggio2013evasion,papernot2016limitations} and the extremely high success rate of attacks, many defense methods have been proposed to defend against these attacks methods~\cite{guo2017countering,buckman2018thermometer,dhillon2018stochastic}.  

%DNNs have been shown to be susceptible to adversarial attacks. An adversarial attack adds special designed perturbations to the original inputs in order to cause DNNs to make mistakes. Due to the wide variety of methods currently used to attack machine learning models~\cite{huang2011adversarial,biggio2013evasion,papernot2016limitations} and the extremely high success rate of attacks, people are increasingly concerned with the safety of machine learning models. 
%Many defense methods that can help machine learning models defend against these attack methods have been proposed in recent years~\cite{buckman2018thermometer,guo2017countering,dhillon2018stochastic}.  

Recently, many efforts have been made to improve the robustness of DNNs, such as  
%Existing defense methods can be divided into two categories: 
(i) using the properties of obfuscated gradients~\cite{athalye2018obfuscated} to prevent the attackers from obtaining the true gradient of the model, e.g., mitigating through randomization~\cite{xie2017mitigating}, Thermometer encoding~\cite{buckman2018thermometer}, and Defense-GAN~\cite{samangouei2018defense}; 
(ii) adding adversarial examples into the training set, e.g., Adversarial Training~\cite{szegedy2013intriguing,goodfellow2014explaining}, scalable Adversarial Training~\cite{kurakin2016adversariala}, and Ensemble Adversarial Training~\cite{tramer2017ensemble}. 
However, it was shown that the first type of defense methods had been broken through by various targeted countermeasures~\cite{DBLP:journals/corr/CarliniW17,he2017adversarial,athalye2018obfuscated}. The second type of methods also suffers the distortion of the classification boundary for the reason that they only import adversarial examples against some specific types of attacks. 
 In this paper, we follow the framework of Adversarial Training and introduce Triplet Loss~\cite{schroff2015facenet}, one of the most popular Distance Metric Learning methods, to improve the robustness by smoothing the classification boundary. Triplet loss is designed to optimize the embedding space such that data points with the same label are closer to each other than those with different labels.  
The primary challenge of triplet loss is how to select representative triplets, which are made up of three examples from two different classes and jointly constitute a positive pair and a negative pair. Since adversarial examples contain more information about the decision boundary than normal examples, we modify the anchor of triplet loss with adversarial examples to enlarge the distance between adversarial examples and examples with different labels in the embedding space. 
Then, we add this fine-grained triplet loss to the original adversarial training process and name the new algorithm as Adversarial Training with Triplet Loss (AT$^2$L).  We also propose an ensemble algorithm which aggregates different types of attacks and model structures to improve the performance. 
Furthermore, the proposed triplet loss can be applied to other methods as a regularization term for better robustness. %For example, most defense methods can use this regularization to further enhance their defense effects. 

We summarize our main contributions as follows:
\begin{itemize}
\item We introduce triplet loss into the adversarial training framework and modify the anchor of triplet loss with adversarial examples. We also design an ensemble version of our method. %We name our proposed algorithm as Adversarial Training with Triplet Loss (AT$^2$L) and also design an ensemble version of AT$^2$L.%We then verify that the new loss function can increase the robustness of model better than the traditional adversarial training process.
%\item We propose to treat the triplet loss as a regularization term.% and verify that our special triplet loss can be no worse than a traditional adversarial training process and effectively increase the robustness of the model.
\item We propose to take our triplet loss as a regularization term and apply it to existing defense methods for further improvement of robustness. % and empirically show that our triplet loss can further improve the defense ability based on current defense methods. 
\item We conduct extensive experiments to evaluate our algorithms. The empirical results show that our proposed approach behaves more robust and preserves the accuracy of the model, and the triplet loss can also improve the performance of other defense methods.

\end{itemize}

\section{Related work}
In this section, we briefly review existing adversarial attack and defense methods. %Some of the attack methods will be used for generating adversarial examples to augment the training set for adversarial training and testing the performance of our defense algorithm. We will also combine our triplet loss and the defense methods which are introduced in this part for the increasing robustness of the model. 

\subsection{Attack methods}
Attack methods can be divided into two main categories: gradient-based attack and optimization-based attack. 

The gradient-based attack asks for the structure of the attacked model and requires that the attacked model should be differentiable. Then it generates adversarial examples by adding perturbation along the direction of the gradients. FGSM~\cite{goodfellow2014explaining}, Single-Step Least-Likely (LL)~\cite{kurakin2016adversarialb,kurakin2016adversariala} and their iterative versions, i.e., I-FGSM and I-LL, are popular methods in this type of attack.

The optimization-based attack formulates the task of attack as an optimization problem which aims to minimize the norm of perturbation and make the DNN model mis-classify adversarial examples. %Therefore,  adversarial examples generated by this attack are more aggressive. 
C\&W attack~\cite{carlini2017towards} is by far one of the strongest optimization-based attacks. It can reduce the classifiers' accuracy to almost 0 and has bypassed over $10$ different methods designed for detecting adversarial examples~\cite{DBLP:journals/corr/CarliniW17}. However, it is more time-consuming than gradient-based algorithms.

\subsection{Defense methods}
Many recent defense approaches are based on a technique called obfuscated gradients~\cite{athalye2018obfuscated}. It is similar to gradient masking ~\cite{papernot2017practical} which is a failed defense method that tries to deny the attacker access to a useful gradient, and leads to a false sense of security in defenses against adversarial examples. Typical defense methods using obfuscated gradients are thermometer encoding~\cite{buckman2018thermometer}, Stochastic activation pruning~\cite{dhillon2018stochastic}, Mitigating through randomization~\cite{xie2017mitigating} and Defense-GAN~\cite{samangouei2018defense}.

%For example, thermometer encoding~\cite{buckman2018thermometer}, Mitigating through randomization~\cite{xie2017mitigating} and Defense-GAN~\cite{samangouei2018defense} are typical methods which apply obfuscated gradients in the defense process. 

Another common method is adversarial training, which proposes to add adversarial examples to the training set and then retrain the model for better robustness.~\citeauthor{szegedy2013intriguing} (\citeyear{szegedy2013intriguing}) first propose this simple process in which the model is trained on adversarial examples until it learns to classify them correctly. However, this type of methods suffers the distortion of the classification boundary. So in this paper, we introduce Distance Metric Learning to alleviate this distortion.

\section{Methodology}
In this section, we first introduce the triplet loss.  
Then we present Adversarial Training with Triplet Loss (AT$^2$L) and an ensemble version of AT$^2$L. %, which integrates multiple attack methods and model structures. 
%Then, we apply this triplet loss to the traditional adversarial training framework. Next, we integrate multiple model structures and attack methods during training to increase the robustness of the model. 
Finally, we propose to treat our special triplet loss as a regularization term and combine it with existing defense methods.
\begin{comment}
\subsection{Notation}
We consider a neural network $f(\cdot)$ used for classification and $f(\mathbf{x})$ represents a probability vector of the prediction. %We classify images, represented as $\mathbf{x}^{w,h,d}$ for a $d$-color image of width $w$ and height $h$.
$\mathbf{x}_i^{adv}$ represents the adversarial example of $\mathbf{x}_i$, and we denote the adversarial examples of $X$ against model structure $m$ under the attack method $a$ as $X^{a,m}$. 
$\|\mathbf{x}_i-\mathbf{x}_j\|$ represents a metric of distance between $\mathbf{x}_i$ and $\mathbf{x}_j$. Here we used $\ell_2$ norm or $\ell_\infty$ as usual. %We denote $\mathbf{x}_i^p$ as the example with the same label as $\mathbf{x}_i^{adv}$, and $\mathbf{x}_i^n$ as the example with different label as $\mathbf{x}_i^{adv}$.
We also quote the definition of triplet loss~\cite{schroff2015facenet}: we denote $\{(\mathbf{x}_i^a,\mathbf{x}_i^p,\mathbf{x}_i^n)\}$ as a triplet, where $(\mathbf{x}_i^a,\mathbf{x}_i^p)$ has the same  label and $(\mathbf{x}_i^a,\mathbf{x}_i^n)$ has  the different. The $\mathbf{x}_i^a$ term is referred to as an anchor of a triplet.

\end{comment}

\subsection{Triplet loss}
%Triplet loss~\cite{schroff2015facenet} first defines triplet data $\{(\mathbf{x}_i^a,\mathbf{x}_i^p,\mathbf{x}_i^n)\}$, where $(\mathbf{x}_i^a,\mathbf{x}_i^p)$ has the same  label and $(\mathbf{x}_i^a,\mathbf{x}_i^n)$ has  the different. The $\mathbf{x}_i^a$ term is referred to as an anchor of a triplet. 
A triplet~\cite{schroff2015facenet} consists of three examples from two different classes, which jointly constitute a positive pair and a negative pair. We denote $(\mathbf{x}_i^a,\mathbf{x}_i^p,\mathbf{x}_i^n)$ as a triplet, where $(\mathbf{x}_i^a,\mathbf{x}_i^p)$ has the same label and $(\mathbf{x}_i^a,\mathbf{x}_i^n)$ has the different. The $\mathbf{x}_i^a$ term is referred to as the anchor of a triplet. The distance between the positive pair is encouraged to be smaller than that of the negative pair, and a soft nearest neighbor classification margin is maximized by optimizing a hinge loss. 
Specifically, triplet loss forces the network to generate an embedding where the distance between $\mathbf{x}_i^a$ and $\mathbf{x}_i^n$ is larger than the distance between $\mathbf{x}_i^a$ and $\mathbf{x}_i^p$ plus the margin parameter $\alpha$.

Formally, we define the triplet loss function as follows:
%The triplet loss function is formulated as follows:
\begin{equation*}
\begin{split}
%\ell(\mathbf{x}) = & 
\frac{1}{N}\sum_{i=1}^N\max\Bigg\{&\|f(\mathbf{x}_i^a)-f(\mathbf{x}_i^p)\| \\
&-\|f(\mathbf{x}_i^a)-f(\mathbf{x}_i^n)\|+\alpha,0\Bigg\},
\end{split}
\end{equation*}
where $N$ is the cardinality of the set of triplets used in the training process, $f(\cdot)$ is the output of the last fully connected layer of our neural network,  $\|\mathbf{x}_i-\mathbf{x}_j\|$ represents a metric of distance between $\mathbf{x}_i$ and $\mathbf{x}_j$. Here we use $\ell_\infty$ norm in our experiments.

%It can also be formulated as 
%\begin{equation*}
%\begin{split}
%\hat \ell(\mathbf{x}) = & \ell(\mathbf{x})+ \frac{1}{N}\sum_{i}\|f(\mathbf{x}_i)-f(\mathbf{x}_i^p)\| \\
%&-\|f(\mathbf{x}_i)-f(\mathbf{x}_i^n)\|+\alpha
%\end{split}
%\end{equation*}
%where $N$ is the number of the triplet data. 
Generating all possible triplets would result in redundant triplets and lead to slow convergence. So in the next sections, we use sampling strategy to generate triplets in our algorithms.
%In this paper, we utilize sampling strategy to generate triplets in our algorithms for the reason that generating all possible triplets would result in redundant triplets, which will not benefit the training process and result in slower convergence. %However, it is crucial to select hard triplets,  which means that, given $\mathbf{x}_i^a$, it is infeasible to select $\mathbf{x}_i^p$ such that $\arg\max_{\mathbf{x}_i^p}\|f(\mathbf{x}_i^a) - f(\mathbf{x}_i^p)\|$ and similarly $\mathbf{x}_i^n$ such that $\arg\min_{\mathbf{x}_i^n}\|f(\mathbf{x}_i^a) - f(\mathbf{x}_i^n)\|$. 	
%So we use sampling strategy to generate triplets in our algorithms.

\subsection{Adversarial training with triplet loss (A$\text{T}^2$L)}
    The original version of adversarial training is to craft adversarial examples for the entire training set and add them to the training process.  Specifically, it generates $X_{adv}$ which contains adversarial examples of instances in training set $X$. Then it concatenates  $X_{adv}$ and $X$ as $X'$ and retrains the model with $X'$. %This is an off-line training process. 
    During each iteration of the original algorithm, it generates the adversarial examples against the current model.  
   The loss function of the original adversarial training is formulated as:
        \begin{equation}
    \begin{split}
         \frac{1}{(1+\lambda)k} \Bigg( \sum_{i=1}^k\ell(\mathbf{x}_i,y_i)+\lambda\sum_{i=1}^k\ell(\mathbf{x}_i^{adv},y_i) \Bigg),
    \end{split}
    \end{equation}
    where $\lambda$ is the hyper-parameter, $k$ is the size of the mini-batch sampled from $X$, $y$ is the label of $\mathbf{x}_i$ and $\mathbf{x}_i^{adv}$ is the adversarial example of $\mathbf{x}_i$.
    
  %The first part of this loss function is a traditional adversarial training process. The second part is a modified triplet loss, which is designed to make the distance between adversarial examples and its original examples smaller than that between adversarial examples and examples with a different label from the original examples. The main difference between this loss and the traditional triplet loss function is that instead of taking the original examples $\mathbf{x}_i$ as the anchor, we choose adversarial example $\mathbf{x}_i^{adv}$, which contains more information about the decision boundary. Our main purpose is to enlarge the margin between different classes and the adversarial examples may have more impact on this purpose. This new loss function will lead to better performance against adversarial attacks.

To encourage a larger margin between the positive class and the negative class, we incorporate triplet loss into the loss function. Specifically, for example $\mathbf{x}_i$, we generate adversarial example  $\mathbf{x}_i^{adv}$ and sampled an example $\mathbf{x}_i^n$ from the mini-batch which has a different label to construct a new triplet $(\mathbf{x}_i^{adv}, \mathbf{x}_i, \mathbf{x}_i^n)$. The main difference between this triplet and the original triplet is that instead of taking the original example $\mathbf{x}_i$ as the anchor, we choose the adversarial example $\mathbf{x}_i^{adv}$, which contains more information about the decision boundary. Specifically, when dealing with a binary classification problem, we sample $\mathbf{x}_i^n$ which has the opposite label to $\mathbf{x}_i$. For multi-class problems, we sample $\mathbf{x}_i^n$ from the same class as the adversarial example $\mathbf{x}_i^{adv}$, which is an incorrect class from the view of human beings.  We apply this new triplet to the triplet loss, and combine it with the loss of adversarial training, so the loss function of our algorithm is formulated as:
% we can define the loss function as 
%    \begin{equation*}
%    \begin{split}
%    	\hat\ell(\mathbf{x},y) = &\frac{1}{m+\lambda k} \Bigg( \sum_{i\in m}\ell(\mathbf{x}_i,y)+\lambda\sum_{i\in k}\ell(\mathbf{x}_i^{adv},y) \Bigg) \\
%        &+ \frac{1}{k}\sum_{i}\max\Bigg\{\|f(\mathbf{x}_i^{adv})-f(\mathbf{x}_i^p)\| \\
%        &-\|f(\mathbf{x}_i^{adv})-f(\mathbf{x}_i^n)\|+\alpha,0\Bigg\}
%    \end{split}
%    \end{equation*}
%  we further generate an adversarial example $\mathbf{x}_i^{adv}$ as $\mathbf{x}_i^a$ in triplet loss, and set original example $\mathbf{x}_i$ as $\mathbf{x}_i^p$ in triplet loss because $\mathbf{x}_i^{adv}$ and $\mathbf{x}_i$ should be classified as the same class from the view of human beings. $\mathbf{x}_i^n$ in triplet loss is sampled from the mini-batch which is classified as different labels from $\mathbf{x}_i$. The proposed triplet loss function can be formulated as: 
    \begin{equation}
    \begin{split}
    	\hat\ell(\mathbf{x},y) =& \frac{1}{(1+\lambda_1)k} \Bigg( \sum_{i=1}^k\ell(\mathbf{x}_i,y_i)+\lambda_1\sum_{i=1}^k\ell(\mathbf{x}_i^{adv},y_i) \Bigg) \\
        &+ \frac{\lambda_2}{k}\sum_{i=1}^k\max\Bigg\{\|f(\mathbf{x}_i^{adv})-f(\mathbf{x}_i)\| \\
        &-\|f(\mathbf{x}_i^{adv})-f(\mathbf{x}_i^n)\|+\alpha,0\Bigg\},
    \end{split}
    \end{equation}
    where $k$ is the size of a mini-batch, and $\lambda_1$, $\lambda_2$ and $\alpha$ are the hyper-parameters. We utilize this new loss function to retrain the model and summarize the proposed algorithm in Algorithm 1. 
    \begin{algorithm}[!htbp]
\caption{Adversarial training with triplet loss (A$\text{T}^2$L)}
\begin{algorithmic}[1]
\STATE Train $f(\cdot)$ with training data $X$;
\REPEAT
\STATE Construct $X_{adv}$ against $f(\cdot)$ for each instance in $X$; 
\STATE $X'$ = [$X$,$X_{adv}$];
\STATE Retrain  $f(\cdot)$ with $X'$ using Eq.~(2);
\UNTIL{Training converged}
\end{algorithmic}
\end{algorithm}

\begin{algorithm*}[!htbp]
\caption{Ensemble version of A$\text{T}^2$L} %of neural network $f(\cdot)$. $A$ is the set of attack methods for training, $M$ is the set of model structures for training}
\begin{algorithmic}[1]
\STATE Train $f(\cdot)$ with training data $X$;
\REPEAT
\FOR {$a$ in $A$}
\FOR {$m$ in $M$}
%\STATE Use attack method $a$ to generate $X^{a,m}$ against model $m$ for each instance in $X$;
\STATE Construct  $X^{a,m}$, which is the set of adversarial examples of $X$ against model $m$ under the attack method $a$.
\ENDFOR 
%\STATE $X_{adv}^{a} = [X^{a,m}]$ for $m$ in $M$
\ENDFOR
\STATE $X_{adv}$ = \{$X^{a,m}$\}, $a \in A$, $m \in M$;
\REPEAT
	\STATE Sample $k$ clean examples $B = \{\mathbf{x}_1 , . . . , \mathbf{x}_k\}$ from training set $X$;
    \STATE Sample $k$ adversarial examples $\{\mathbf{x}_1^{adv}, ... , \mathbf{x}_k^{adv}\}$ %from corresponding clean examples $\{\mathbf{x}_1 , ... , \mathbf{x}_k\}$, each $\mathbf{x}_i^{adv}, i=1,2,...,k$ is sampled 
    from $X_{adv}$. Each $\mathbf{x}_i^{adv}$ is the adversarial example of $\mathbf{x}_i$;   
    \STATE Construct a new training batch $B' = \{\mathbf{x}_1 , . . . , \mathbf{x}_k, \mathbf{x}_1^{adv},$ $ ... , \mathbf{x}_k^{adv}\}$;
    \STATE For each instance of $\{\mathbf{x}_1^{adv}, ... , \mathbf{x}_k^{adv}\}$, take $\mathbf{x}_i$ as $\mathbf{x}_i^p$ in triplet loss and sample an example from $B$ with a different label from $\mathbf{x}_i$ as $\mathbf{x}_i^n$ in triplet loss;
    \STATE Perform one training step of network $f(\cdot)$ using the mini-batch $B'$ according to Eq.~(2);
\UNTIL {Training converged}
\UNTIL{Training converged}
\end{algorithmic}
\end{algorithm*}
  
\subsection{Ensemble A$\text{T}^2$L}
	We proceed to improve the robustness of the model against unknown type of attacks for the reason that the originally proposed algorithm can only defend against known type of attacks, where defenders have detailed information about the attacking methods and lack robustness against attacks transferred from unknown models.
	%The algorithm mentioned above only cares about the margin between different classes, the process of generating adversarial examples is also very important. The simple way is to generate adversarial examples  by single attack method, e.g. FGSM or C\&W. 
    Our first attempt is to combine different attack methods together to increase the robustness. % So we propose an ensemble version of A$\text{T}^2$L which is shown in Algorithm 2.
	%Here we proposed three versions of adversarial training.
    As shown in Algorithm 2 where $A$ denotes an aggregation of attack methods, we conduct adversarial training on a collection of adversarial examples that are generated by all the attack methods. In this paper, we consider three types of attacks as follows: 
    \begin{itemize}
    \item Gradient-based: $A=\{\text{FGSM}, \text{LL}, \text{I-FGSM}, \text{I-LL}\}$. 

    \item Optimization-based: 
    	$A=\{\text{C\&W}\}$.

    \item Mixed: $A=\{ \text{FGSM}, \text{LL}, \text{I-FGSM}, \text{I-LL}, \text{C\&W}\}$.
    
    \end{itemize}
    
    On the other hand, we adopt the idea of Ensemble Adversarial Training~\cite{tramer2017ensemble}, which says that the augmentation of training data with perturbations transferred from other models can improve the robustness not only under a known type of attack, but also under an unknown type of attack. As shown in Algorithm 2, where $M$ is a set of model structures, we extend our training set with adversarial examples against different models in $M$.

    %When optimizing our triplet loss with batch-SGD, we sampled each batch from this large $X_{adv}$. 
   %For better robustness against attacks under both white-box and black-box settings,  we  extend our algorithm with different model structures. As shown in Algorithm 2, $M$ is a set of models with different structures.~\citeauthor{tramer2017ensemble} (\citeyear{tramer2017ensemble}) have proved that a future black-box adversary can not be stronger than the coalition of some static distributions. So we generate adversarial examples against different model structures and these adversarial examples can be seen as being sampled from different distributions. 
 %Intuitively, as adversarial examples transfer between models with different structures, perturbations crafted on an external model are good approximations for solving the min-max formulation of adversarial training, which is described in Formulation (1). This augmentation of training data with perturbations transferred from other models can improve the robustness not only under a white-box setting, but also under a black-box setting. 

In general, the ensemble version of our algorithm not only considers various types of attacks, but also involves adversarial examples generated against different model structures. Therefore, our algorithm captures more information about the decision boundary, and with our designed triplet loss, it can smooth the classification boundary and learn a better embedding space to alleviate the distortion.

\begin{figure*}[!htb] 
\subfigure[Cats vs. Dogs]{
\begin{minipage}[b]{0.33\textwidth}
\centering
\includegraphics[width=6cm]{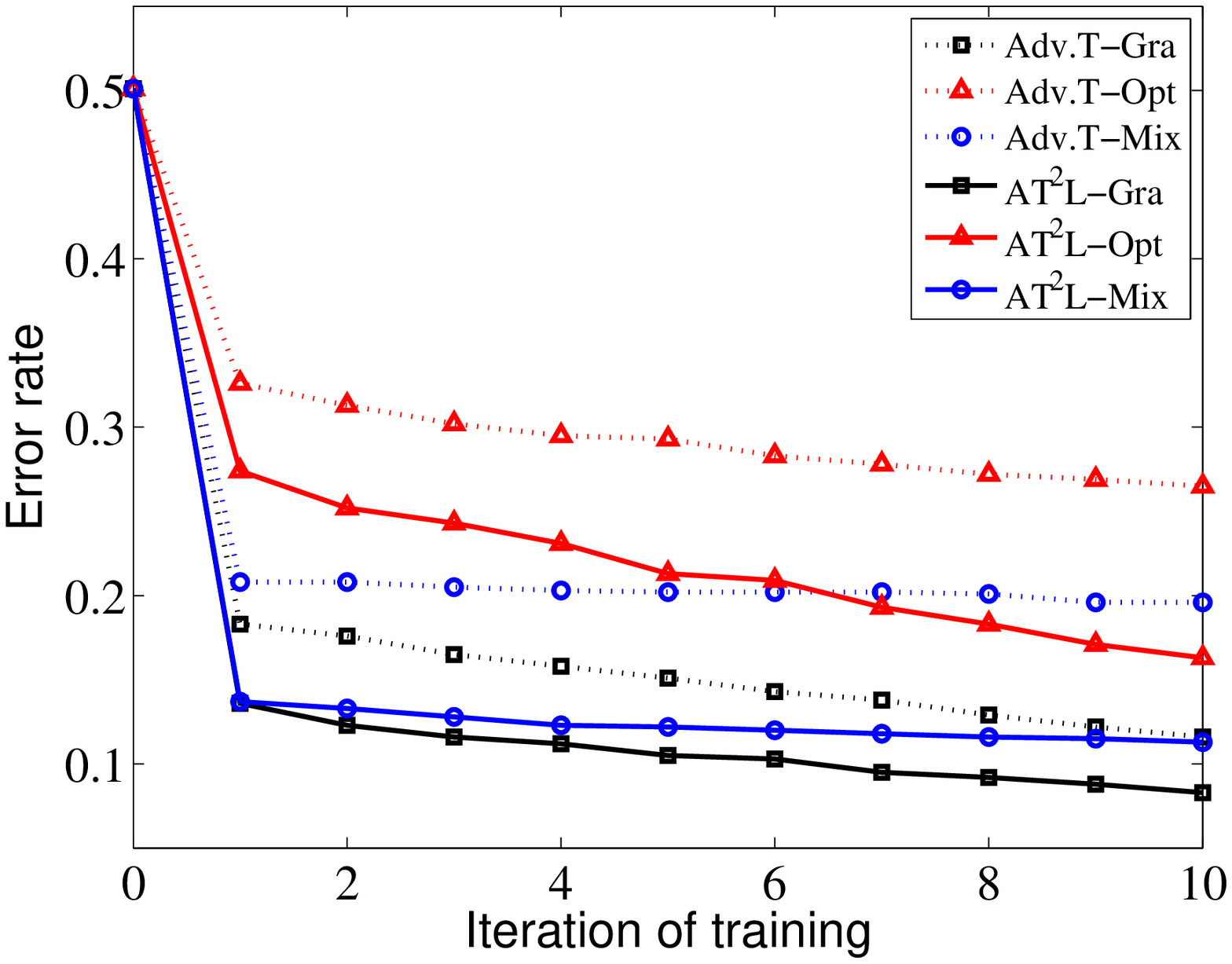}
\includegraphics[width=6cm]{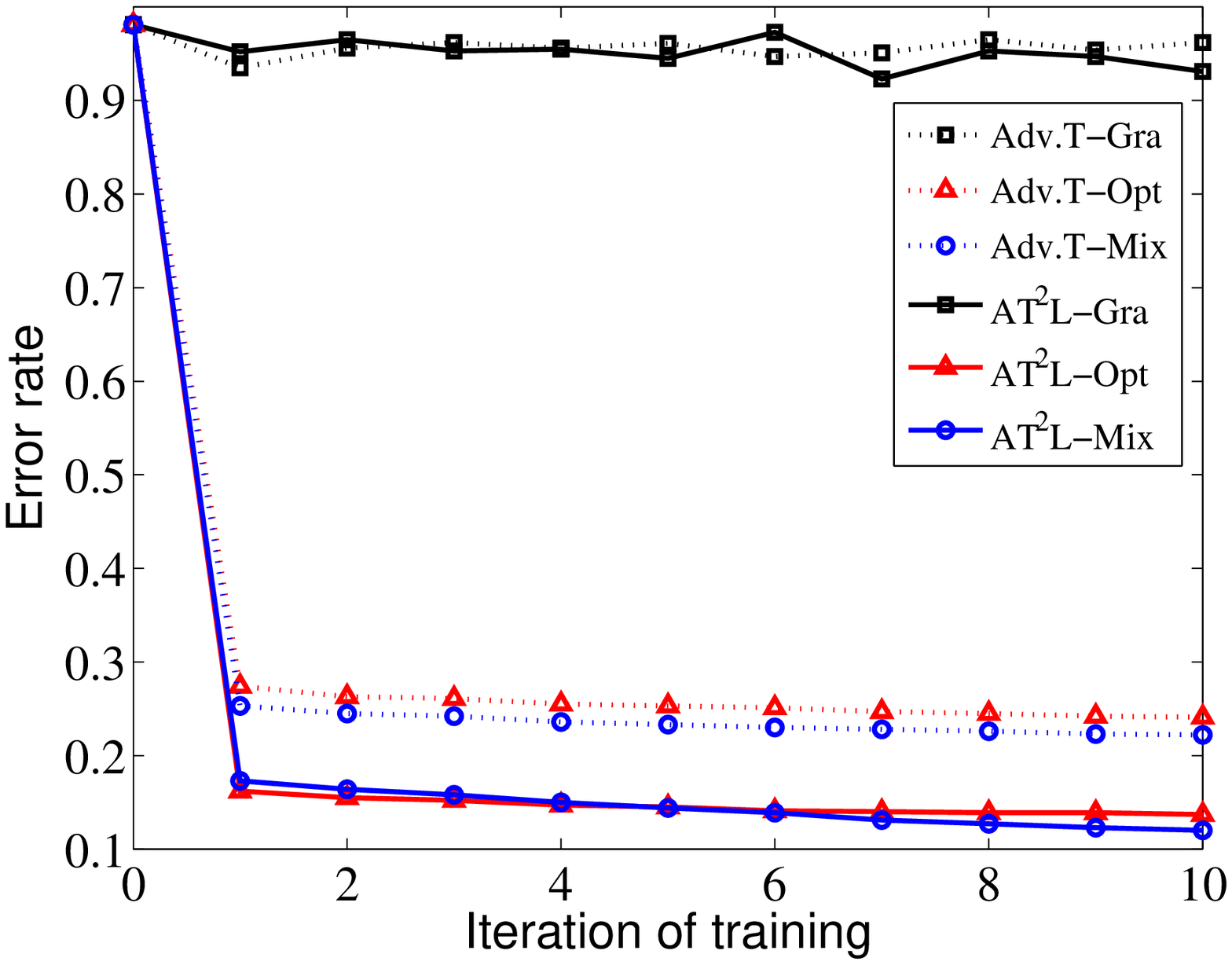}
\end{minipage}
}
\subfigure[MNIST]{
\begin{minipage}[b]{0.33\textwidth}
\centering
\includegraphics[width=6cm]{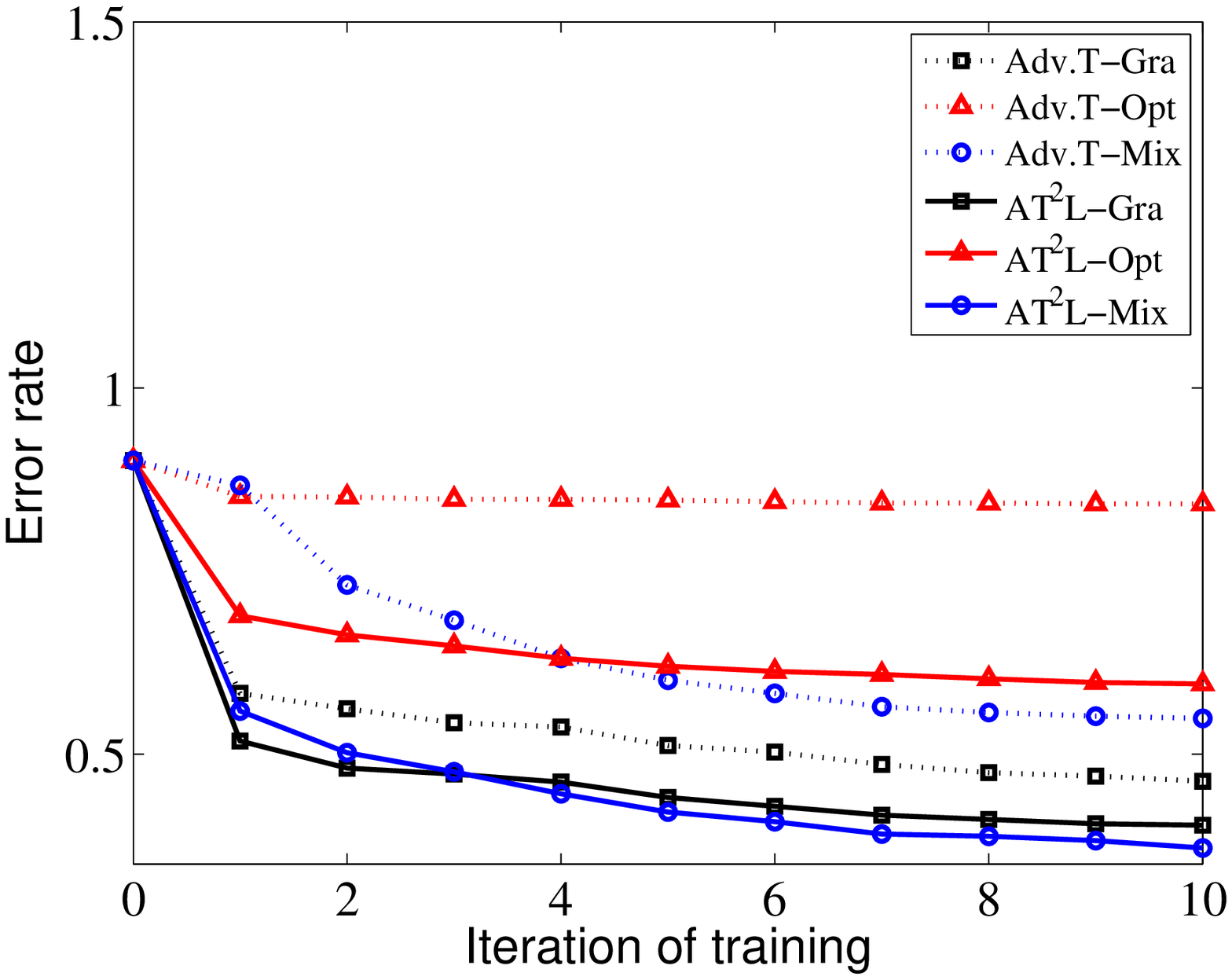}
\includegraphics[width=6cm]{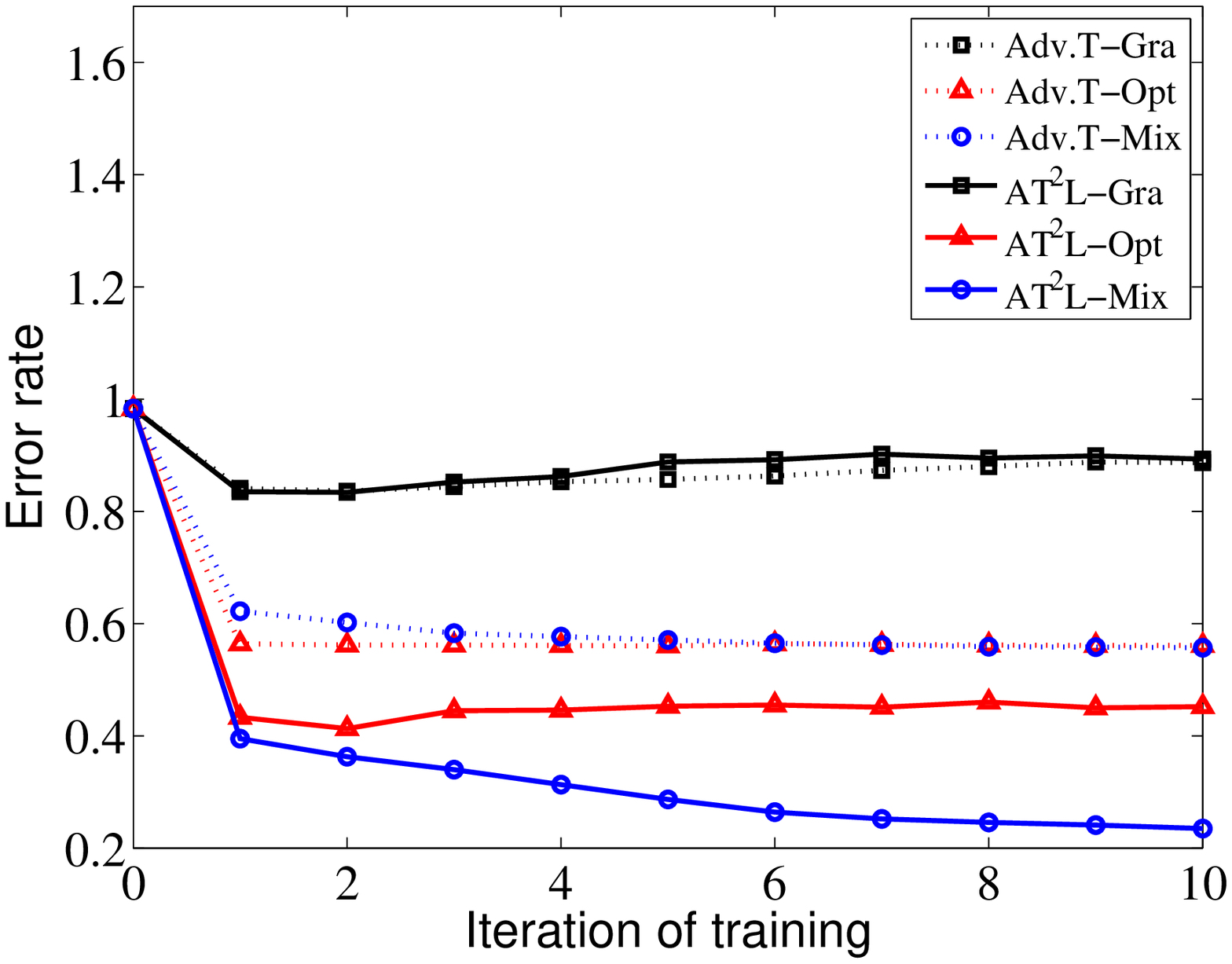}
\end{minipage}
}
\subfigure[CIFAR10]{
\begin{minipage}[b]{0.33\textwidth}
\centering
\includegraphics[width=6cm]{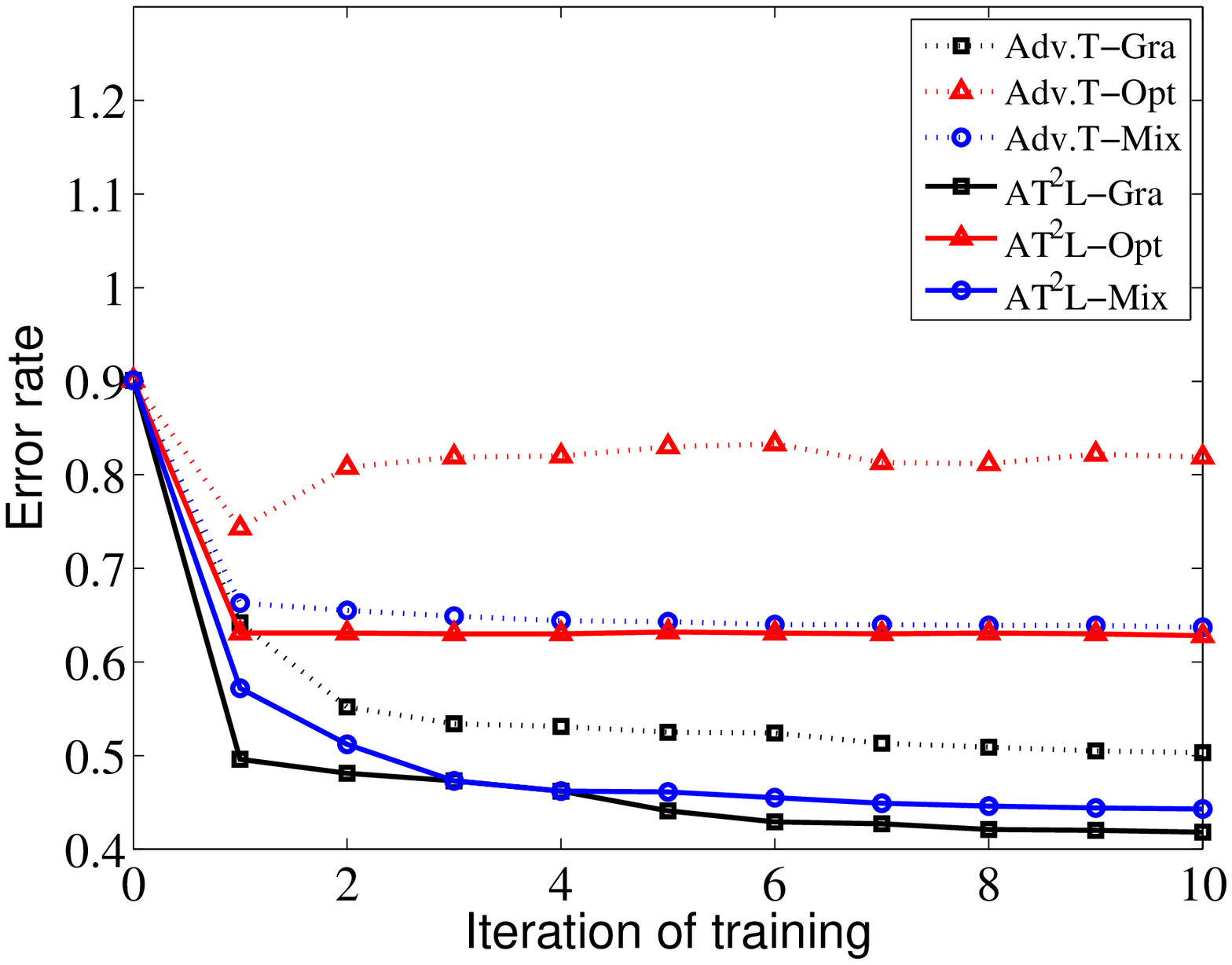}
\includegraphics[width=6cm]{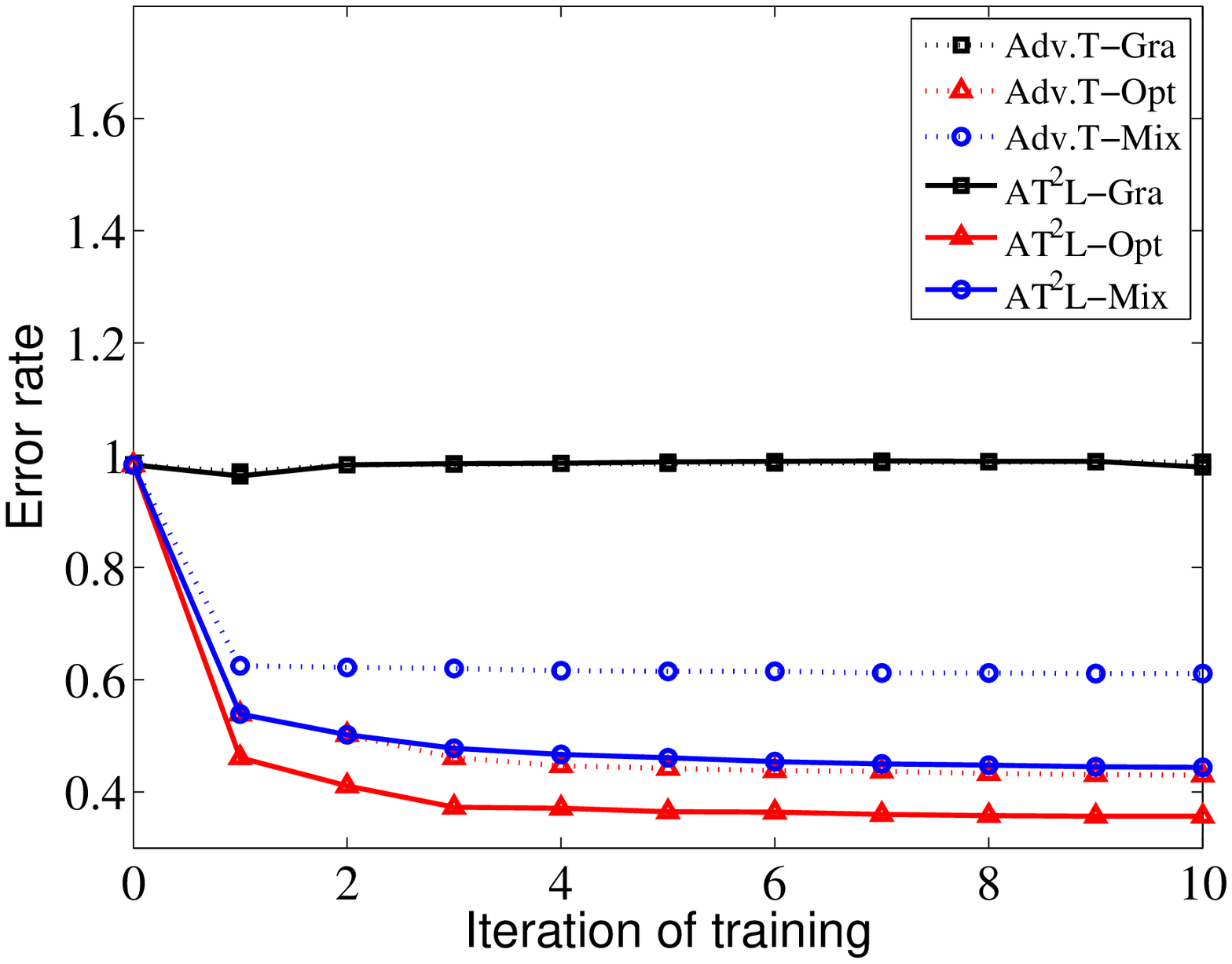}

\end{minipage}
}
\caption{Results on three datasets where attackers perform white-box attacks. `Adv.T' means the traditional adversarial training. `-Gra' means the training process uses the gradient-based attack methods to generate adversarial examples. `-Opt' means using optimization-based attack method, i.e., C\&W and `-Mix' means using the mixed version of attack methods. The figures of the upper line are attacked by FGSM and figures of the bottom line are attacked by C\&W.}
\label{fig:z2} 
\end{figure*} 

\begin{figure*}[h] 
\centering
\subfigure[Cats vs. Dogs]{
\begin{minipage}[b]{0.31\textwidth}
\centering
\includegraphics[width=6cm]{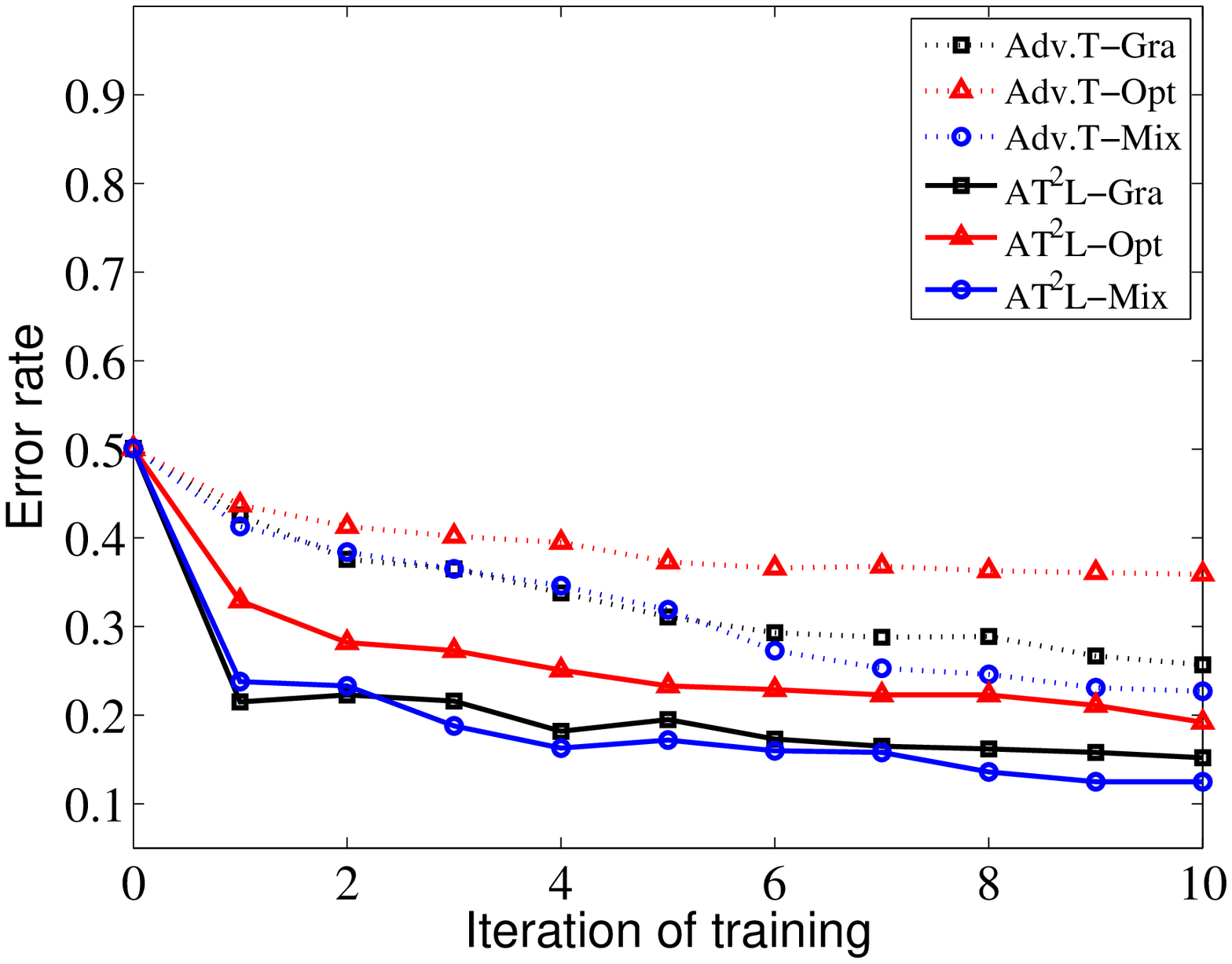}
\end{minipage}
}
\subfigure[MNIST]{
\begin{minipage}[b]{0.31\textwidth}
\centering
\includegraphics[width=6cm]{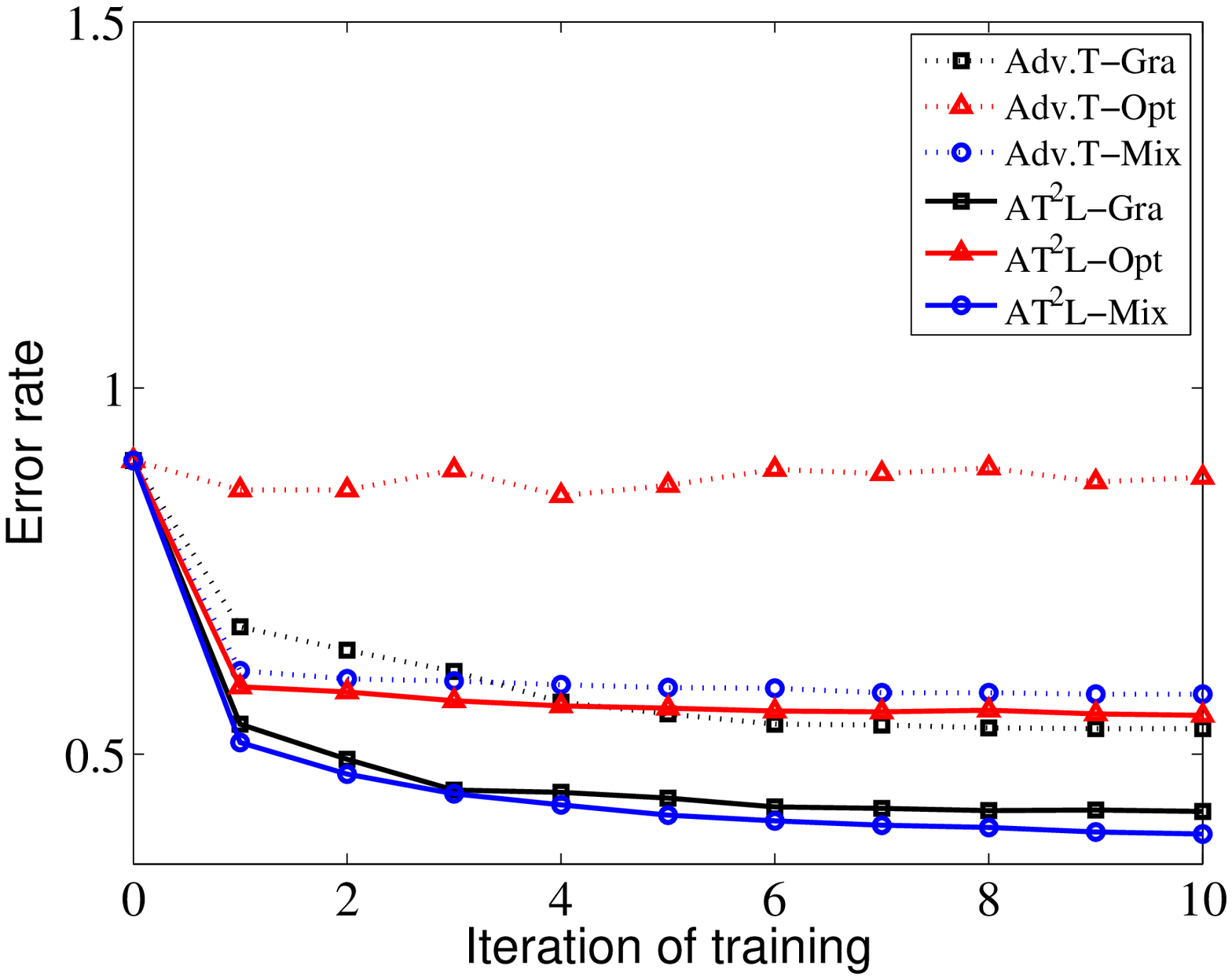}
\end{minipage}
}
\subfigure[CIFAR10]{
\begin{minipage}[b]{0.31\textwidth}
\centering
\includegraphics[width=6cm]{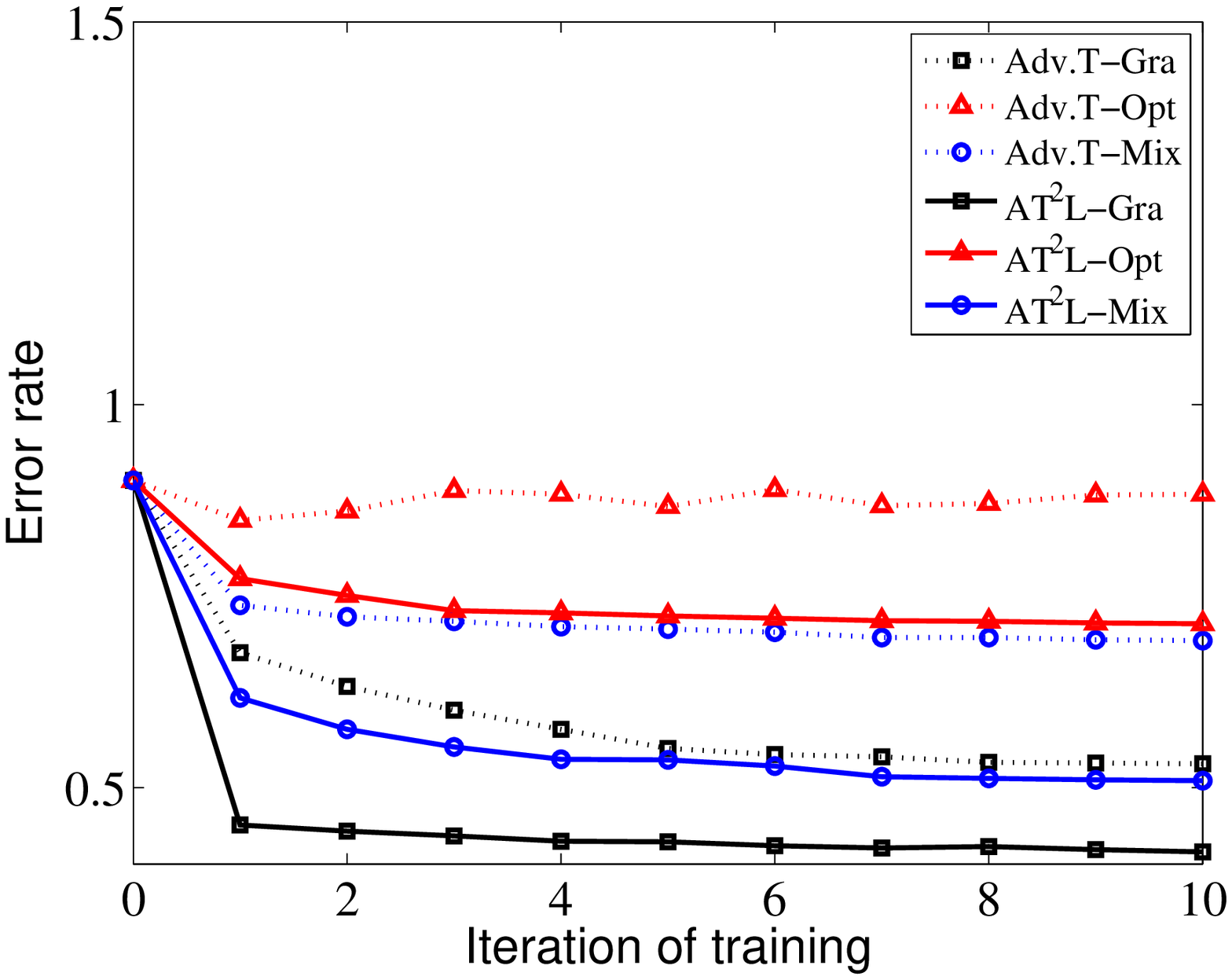}
\end{minipage}
}
\caption{Results on three datasets where defenders are under the attack of adversarial examples transferred from unknown models. Notations are the same as Figure 1 and these  are attacked by FGSM.}
\label{fig:z1} 
\end{figure*} 

\subsection{Triplet regularization}
%When introducing triplet loss into the adversarial training framework, we proposed to enlarge the margin between the distance of different classes. We take adversarial examples other than original examples for the reason that these examples contain more information about the boundary and the distance calculated between these examples can be more representative. 

%The algorithms mentioned above emphasize the adversarial training process. However, the proposed triplet loss can also play such a role to alleviate the distortion and improve the robustness of the model. %not only alleviate the distortion caused by adversarial training but also improve the robustness of the model. 
%We use the adversarial examples as the anchor examples which are more like a wrong class from the view of original model, so that the distance between the anchor and the negative examples is less than that between the anchor and the positive examples. The triplet loss is introduced into the framework of adversarial training for the purpose of fixing the `error'. 
%We use the adversarial examples as the anchor examples which are more like a wrong class from the view of original model. However, from the view of human beings, it is an obvious `error' that the distance between the adversarial example and the negative example is less than that between the adversarial example and the positive example. Our triplet loss can fix this `error' by enlarging the margin between the adversarial examples and the negative examples and decreasing the margin between examples with the same class.  
Our triplet loss can also be regarded as a regularization term:
%The purpose of adversarial training is consistent with our triplet loss, so we propose to explore the effect when trained with our triplet loss without adversarial training. If we consider only our specially designed triplet loss, we may get a new regularization:
        \begin{equation*}
    \begin{split}
    	%\hat\ell(\mathbf{x},y) =& \frac{1}{m}  \sum_{i\in m}\ell(\mathbf{x}_i,y) 
        R(\mathbf{x},y) =& \frac{\lambda}{k}\sum_{i=1}^k\max\Bigg\{\|f(\mathbf{x}_i^{adv})-f(\mathbf{x}_i)\| 
        \\&-\|f(\mathbf{x}_i^{adv})-f(\mathbf{x}_i^n)\|+\alpha,0\Bigg\}
    \end{split}
    \end{equation*}
    
Thus, it can be incorporated into most of the existing defense methods for better robustness. The defense methods based on obfuscated gradients mostly mask the real gradient by adding non-differentiable preprocessing or random processes, and there is no restriction on the loss function used in the training process. Therefore, we can modify their loss function by adding our triplet regularization term to further increase the robustness.
    
    For example,~\citeauthor{buckman2018thermometer} (\citeyear{buckman2018thermometer})  propose to encode the input with Thermometer Encoding and retrain the model with the traditional adversarial training. Triplet regularization can be easily applied to this method by changing the loss function of the adversarial training process. Mitigating through randomization~\cite{xie2017mitigating} and Defense-GAN~\cite{samangouei2018defense} both perform transformations over original inputs without changing the loss. So we can directly incorporate our triplet regularization into their losses to improve the defense effect.

\begin{figure*}[h] 
\centering
\subfigure[Cats vs. Dogs]{
\begin{minipage}[b]{0.31\textwidth}
\centering
\includegraphics[width=6cm]{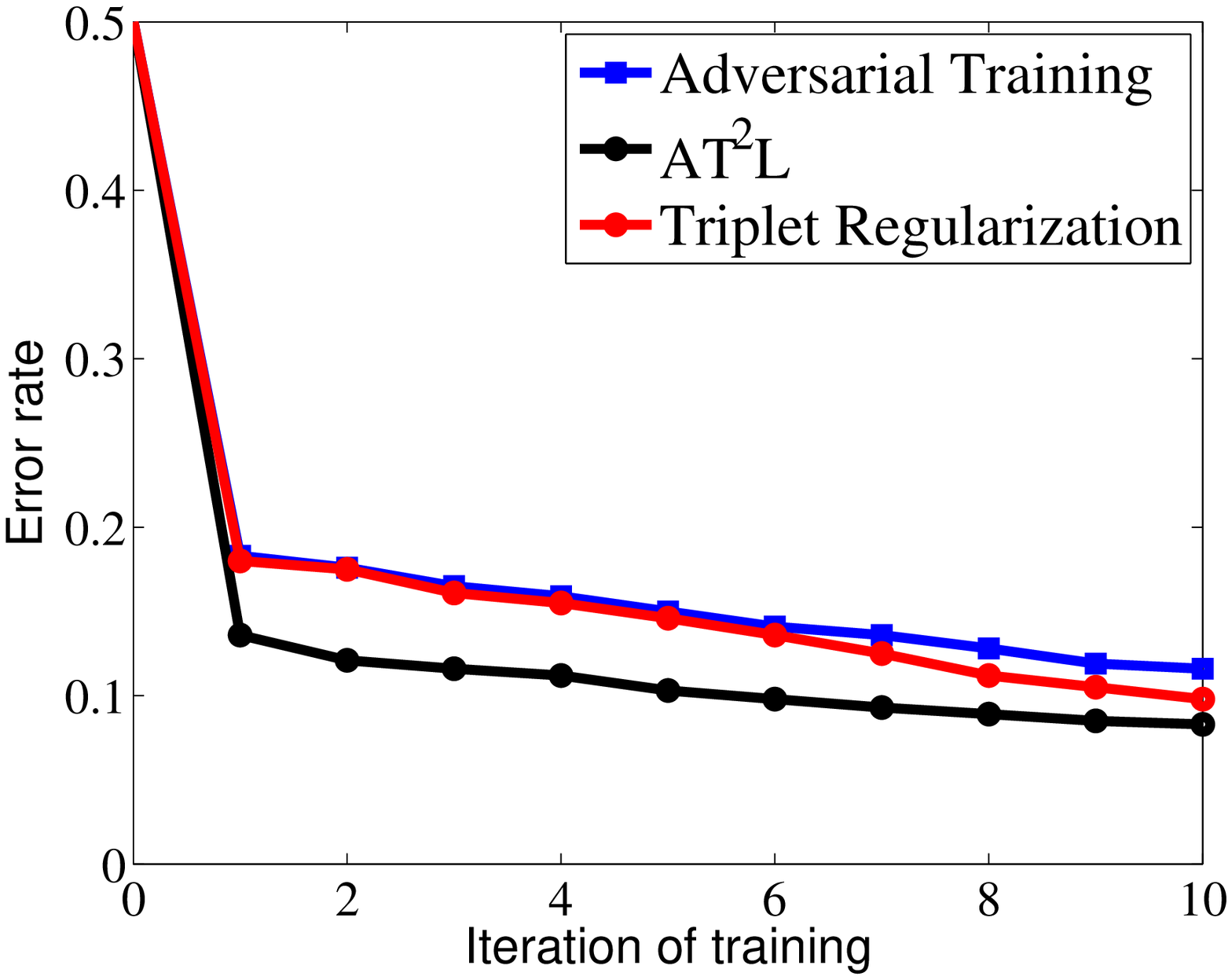}
\end{minipage}
}
\subfigure[MNIST]{
\begin{minipage}[b]{0.31\textwidth}
\centering
\includegraphics[width=6cm]{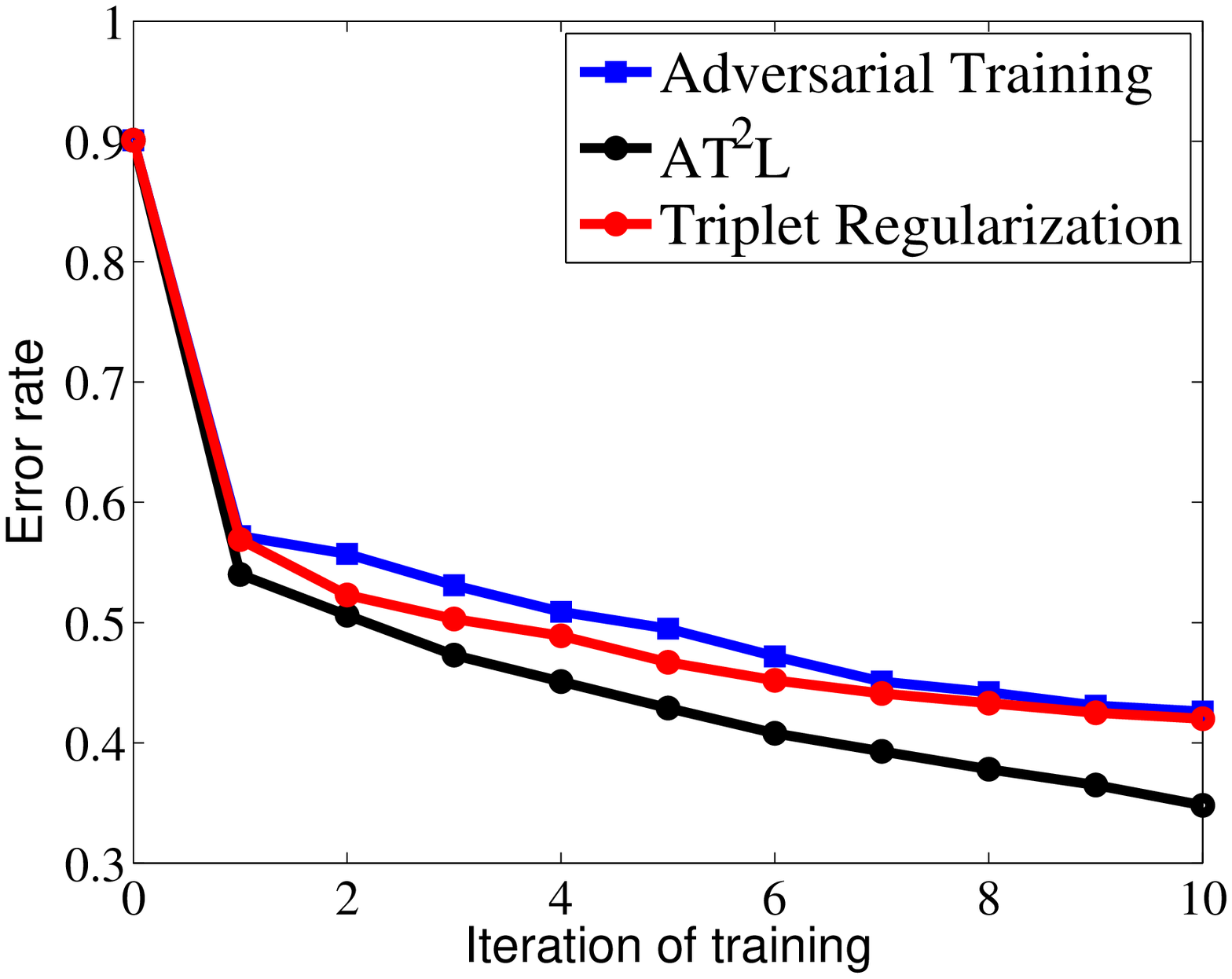}
\end{minipage}
}
\subfigure[CIFAR10]{
\begin{minipage}[b]{0.31\textwidth}
\centering
\includegraphics[width=6cm]{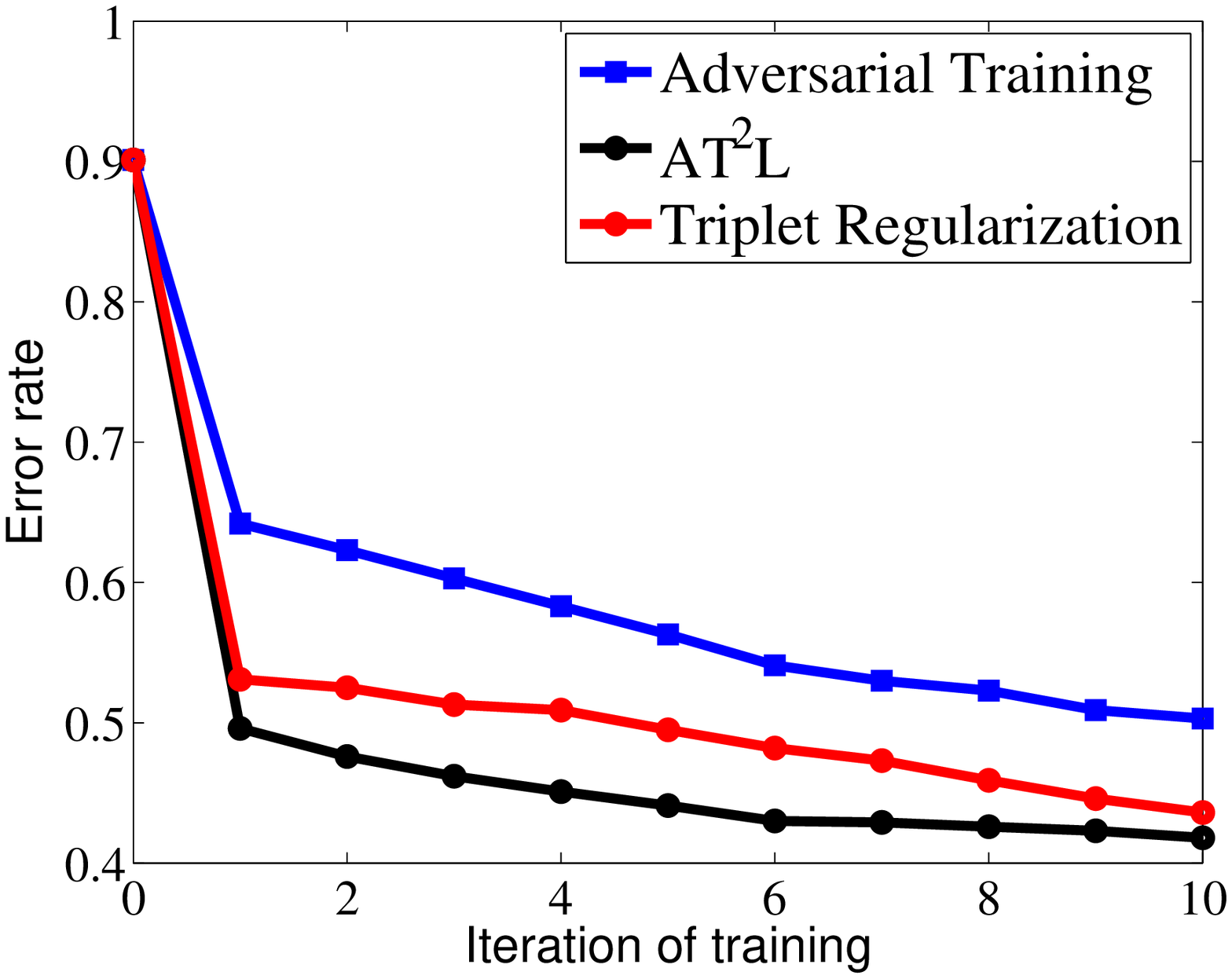}
\end{minipage}
}
\caption{Comparison of  traditional adversarial training and triplet regularization. The attack method used is FGSM.}
\label{fig:z1} 
\end{figure*} 

\section{Experiments}
%In this section, we claim the datasets we used and the setting of our experiments. Then we compare the results of normal adversarial training and our ensemble version of Adversarial Training with Triplet Loss (AT$^2$L) on different datasets. We then test the robustness of the model when only train our model with our special triplet loss as a regularization term. Finally, we do experiments over some current defense methods.% to check that whether our AT$^2$L and triplet regularization can further improve the robustness of the model or not.
In this section, we present experimental results.

\subsection{Settings}
We conduct experiments over three datasets, i.e., Cats vs. Dogs~\cite{elson2007asirra}, MNIST ~\cite{lecun1998mnist} and CIFAR10~\cite{krizhevsky2009learning}. Cats vs. Dogs is a large scale image dataset used for binary classification problems. MNIST and CIFAR10 are commonly used datasets for multi-class classification problems.
The attack methods we used in our experiments include FGSM, I-FGSM, LL, I-LL, C\&W, LS-PGA and Deepfool~\cite{moosavi2016deepfool} and the model structures used in the experiments are different for three datasets. % and more detailed information are  provided in the supplemental material\footnote{\url{https://github.com/Anonymousforijcai/appendix4ijcai}}.
The parameters of these methods, detailed model structures and full results of the experiments are described in the supplemental material\footnote{\url{https://github.com/njulpc/IJCAI19/blob/master/Appendix.pdf}}. %We set $\epsilon$ as $0.3$, which is the scale of FGSM, I-FGSM, LL and I-LL. For iteration algorithms like I-FGSM and I-LL, we set the number of iteration as 10 and $\epsilon = 0.03$ for each iteration. 

\subsection{Adversarial Training with Triplet Loss (A$\text{T}^2$L)}

To illustrate the advantage of the proposed method, we compare it with adversarial training without triplet loss, whose loss function is  Eq.~(1). %This is the traditional adversarial training process which directly introduces adversarial examples into the training set to increase robustness.  
As for the hyper-parameters in  Eq.~(1) and Eq.~(2), we traverse in the appropriate interval and find that they have a stable performance in a proper range of values. 
Each experiment is tested by two types of attack methods, i.e., FGSM and C\&W.  Due to the limitation of space,  we only show results where attackers perform white-box attacks in Fig.~1 and partial results where defenders perform the attack to a network which is not included in the training set $M$ of our algorithm in Fig.~2. 

%From the results, we find that compared with the model trained over clean data, all the models trained with our algorithm, i.e., A$\text{T}^2$L, have no loss of accuracy, and more details can be found in the supplemental material. When the model is under attacked, the error rates of all methods keep decreasing as the number of iterations increases, except when the model trained with gradient-based attacks is attacked by an optimization-based attack. However, our mixed version of algorithm can address this problem. In general, all results illustrate that our  AT$^2$L performs better than the original adversarial training. 

%The model after normal adversarial training or A$\text{T}^2$L does not lose the accuracy  comparing to model trained over clean data, and all results of black-box attack are slightly worse than the white-box attack. %, but the robustness of the model indeed increases. 
%The reason is that black-box attack is lack of precise information of the model, i.e., the type of the layers and the gradient of the model, but we use the property that black-box attacks are easily transferrable from one model to the other and train our model with some similar models so that the robustness of the model can indeed increase.  
%From Figure 1 and Figure 2, we  can see that our  A$\text{T}^2$L shows better performance than  traditional adversarial training trained with corresponding attacks under both FGSM and C\&W attacks.

We have the following observations from the results in Fig.~1: (i) when attacked by gradient-based attacks or optimization-based attacks, A$\text{T}^2$L trained with adversarial examples generated by corresponding attacks has the best robustness, e.g., when attacked by gradient-based attacks, the model trained with adversarial examples generated by gradient-based attacks exhibits the best robustness; (ii) when trained with gradient-based attacks, A$\text{T}^2$L shows almost no robustness against optimization-based attacks, which is shown by the black curves in Fig.~1. However, the robustness of our algorithm trained with optimization-based attacks demonstrates a decent defense effect against gradient-based attacks. This briefly verifies that optimization-based attacks are stronger and contain more information about the decision boundary than gradient-based attacks; (iii) A$\text{T}^2$L trained with our mixed version of algorithms shows comparable robustness to the model trained with corresponding attacks. Although the mixed version of A$\text{T}^2$L does not perform the best, it provides more reliable robustness when attacked by an unknown type of attacks.

Compared with Fig.~1 (results under adversarial examples transferred from known models), Fig.~2  (results under adversarial examples transferred from unknown models) shows that the results under the attack transferred from unknown models are slightly worse than that under the known type of attacks for the reason that the defenders are lack of precise information of the attack, e.g., the type of the attack method and model structure used for attack. However, the model still shows decent robustness against unknown type of  attacks, and this is an advantage of our ensemble  A$\text{T}^2$L, which aggregates multiple model structures.

We also find that compared with the model trained over clean data, all the models trained with our algorithm, i.e., A$\text{T}^2$L, have no loss of accuracy, and more details can be found in the supplemental material.

\begin{table*}[!htbp]
\centering
\label{lz9}
\begin{tabular}{|c|c|c|c|c|c|c|c|c|c|c|c|c|c|c|c|}
\hline
 & \multicolumn{3}{c|}{Ori}  & \multicolumn{3}{c|}{Th. En.(1)} & \multicolumn{3}{c|}{Th. En.(7)} & \multicolumn{3}{c|}{Th. En. + TR(1)} & \multicolumn{3}{c|}{Th. En. + TR(7)} \\ \hline
Clean & \multicolumn{3}{c|}{\textbf{5.8}}  & \multicolumn{3}{c|}{7.6}           & \multicolumn{3}{c|}{10.1}          & \multicolumn{3}{c|}{6.1}                     & \multicolumn{3}{c|}{7.2}                     \\ \hline
FGSM  & \multicolumn{3}{c|}{51.5} & \multicolumn{3}{c|}{37.1}          & \multicolumn{3}{c|}{20.0}          & \multicolumn{3}{c|}{27.6}                    & \multicolumn{3}{c|}{\textbf{15.1}}                    \\ \hline
PGD/LS-PGA  & \multicolumn{3}{c|}{49.5} & \multicolumn{3}{c|}{39.3}          & \multicolumn{3}{c|}{20.9}          & \multicolumn{3}{c|}{29.7}                    & \multicolumn{3}{c|}{\textbf{13.4}}                    \\ \hline
\end{tabular}
\caption{Error rate against known type of attacks on CIFAR10 over Thermometer Models. `Th. En.' mean Thermometer Encoding. `TR' means applying our triplet regularization.}
\end{table*}

\begin{table*}[!htbp]
\centering

\label{lz6}
\scalebox{0.9}[0.9]{
\begin{tabular}{|c|c|c|c|c|c|c|c|c|c|c|c|c|}
\hline
Model  & \multicolumn{3}{c|}{Inception-v3}                                  & \multicolumn{3}{c|}{ResNet-v2-101}                                 & \multicolumn{3}{c|}{Inception-ResNet-v2}                            & \multicolumn{3}{c|}{Ens-adv-Inception-ResNet-v2}                    \\ \hline
         & Ori   & Rand & Rand + TR & Ori   & Rand & Rand + TR & Ori   & Rand & Rand + TR & Ori  & Rand & Rand + TR \\ \hline
FGSM     & 66.8  & 36.2 & \textbf{30.5}                                                & 73.7  & 28.2 & \textbf{21.7}                                                & 34.7  & 19.0 & \textbf{8.3}                                                 & 15.6 & \textbf{4.3}  & 4.6                                                 \\ \hline
Deepfool & 100.0 & 1.7  & \textbf{1.1}                                                 & 100.0 & 2.3  & \textbf{1.5}                                                 & 100.0 & 1.8  & \textbf{0.8}                                                 & 99.8 & 0.9  & \textbf{0.7}                                                 \\ \hline
C\&W     & 100.0 & 3.1  & \textbf{2.6}                                                 & 100.0 & 2.9  & \textbf{1.2}                                                 & 99.7  & 2.3  & \textbf{1.3}                                                 & 99.1 & 1.2  & \textbf{0.9}                                                 \\ \hline
\end{tabular}}
\caption{Error rate of different models under the vanilla attack scenario on the ImageNet datasets. `Ori' means the original model. `Rand' means adding some randomization layers. `TR' means applying our triplet regularization. }
\end{table*}
\begin{table*}[!htbp]
\centering
\scalebox{0.9}[0.9]{
\begin{tabular}{|c|c|c|c|c|c|c|c|c|c|c|c|c|}
\hline
Model             & \multicolumn{3}{c|}{A}                                     & \multicolumn{3}{c|}{B}                                     & \multicolumn{3}{c|}{C}                                     & \multicolumn{3}{c|}{D}                                     \\ \hline

 & {Ori} & {DG} & DG + TR          & {Ori} & {DG} & DG + TR          & {Ori} & {DG} & DG + TR          & {Ori} & {DG} & DG + TR \\ \hline
FGSM              & 88.3                 & 1.2                  & \textbf{1.1} & 97.8                 & 4.4                  & \textbf{0.7} & 67.9                 & 1.1                  & \textbf{0.8} & 96.2                 & 2.0                  & \textbf{1.6} \\ \hline
C\&W              & 85.9                 & \textbf{1.1}         & 1.4          & 96.8                 & 8.4                  & \textbf{4.7} & 87.4                 & \textbf{1.1}         & \textbf{1.1} & 96.8                 & 1.7                  & \textbf{1.4} \\ \hline
\end{tabular}}
\caption{Error rates of different models on the MNIST datasets. `DG' mean Defense-GAN. `TR' means applying our triplet regularization. A,B,C,D are different model structures, whose details are described in the supplemental material.}
\end{table*}
\subsection{Triplet regularization}

%The use of triplet loss is proposed to decrease the margin between different classes. %We use the adversarial examples as the anchor examples which is more like a wrong class from the view of original model so that the distance of the anchor and the negative examples is less than the distance of the anchor and the positive examples. The triplet loss is introduced into the framework of adversarial training for the purpose of fixing the 'error'. 
%The adversarial training part plays the similar role as our triplet loss so we propose to explore the defense effect when only trained with triplet regularization. 

%We set the margin $\alpha$ to be $0.5, 1.0, 1.0$ for Cats vs. Dogs, MNIST and CIFAR10. Parameter $k$ is set to be $32$, and $\lambda$ is set to be $0.1$. 
To reveal the effect of triplet regularization, we compare it with the original adversarial training. % and our AT$^2$L.  
We use FGSM to generate adversarial examples for the training process and test the robustness by the attack of FGSM.

From Fig.~3, we can see that the error rate of the model trained with our triplet regularization (the red curve) keeps decreasing as the number of iterations increases. So the triplet loss itself can increase the robustness of the model, and its effect is no worse than the original adversarial training (the blue curve). %However, our adversarial training with triplet loss still shows the best performance among them.  The result indicates that our special designed triplet loss can increase robustness excluding the traditional adversarial training.  
This result verifies our hypothesis that enlarging the margin between the adversarial examples and the negative examples and decreasing the margin between examples with the same class can smooth the decision boundary. %, and the margin between different classes truly affects the robustness of the model.  
This also suggests that the designed triplet regularization can work well in most machine learning problems to increase robustness.  We can also see form Fig.~3 that A$\text{T}^2$L, which integrates both adversarial training and triplet regularization, shows the best performance. 
% Although our Adversarial Training with Triplet Loss still shows the best performance, our triplet regularization can be easily applied to other defense methods without any extra modification of model structures or training procedure. 

\subsection{Current defense methods with triplet regularization}
We further explore the effect of the combination of triplet regularization with existing defense methods. %Although most defense methods are claimed to be broken through recently~\cite{athalye2018obfuscated}, the robustness of the model under these methods is truly increased, and the attackers need a particular measure to overcome these defenses. So 
We experiment over some representative defense methods and demonstrate that our triplet regularization can be applied to improve their robustness further. %For a fair comparison, the model structures used in each defense are the same as that in their original papers. 
Due to the limitation of space, we show partial results in this part, and full results are listed in the supplemental material.

\subsubsection{Thermometer encoding}

	%\citeauthor{buckman2018thermometer} (\citeyear{buckman2018thermometer}) demonstrated that the use of thermometer encodings, in combination with adversarial training, can reduce the vulnerability of neural network models to adversarial attacks. The original setting of~\citeauthor{buckman2018thermometer}s' work was to encode the input with thermometer encoding and to train using traditional adversarial training. Triplet regularization can be easily applied into this method by changing the loss function of the adversarial training process.
   
We follow the setting of the original paper~\cite{buckman2018thermometer} and do experiments over both known  and unknown type of attacks. %We compare the original model which is trained over clean data, and the model trained with Thermometer encoding and the model combined with triplet regularization. We trained the model up to 7 iterations for further comparison between the defense ability of our triplet regularization and that of the thermometer encoding. 
  Partial results are shown in Table 1 which indicate employing our triplet regularization indeed improves the robustness based on the effect of the original defense. For example, when attacked by FGSM, the error rate of the model trained using thermometer encoding after 7 iterations is 20.0\%. However, combining with our triplet regularization, the model can achieve 15.1\% error rate.

\subsubsection{Mitigating through randomization}

We also apply our triplet loss to the work of~\citeauthor{xie2017mitigating} (\citeyear{xie2017mitigating}) who proposed to randomly resize or pad the images to a designed size. This defense can be added in front of normal classification process with no additional training or fine-tuning, and can be combined with our triplet regularization directly. We experiment over two settings from the original paper (vanilla attack scenario and ensemble-pattern attack scenario), and examine the performance of our triplet regularization. The result of the vanilla attack scenario is shown in Table 2. For the model of Inception-ResNet-v2, the randomized procedure only achieves 19.0\% error rate under FGSM, but with our triplet regularization, the error rate can drop to 9.3\%. These results show that our triplet regularization can further improve the robustness of the model based on the original defense method. 

\subsubsection{Defense-GAN}

Defense-GAN is designed to project samples onto the manifold of the generator before classifying them. %This defense process is also separated from the classification model and the training process of the classifier. 
Our triplet regularization can be easily applied after the projection of Defense-GAN by simply changing the loss function during the training process of the classifier. We follow the setting of models' structures and parameters in the paper of~\citeauthor{samangouei2018defense} (\citeyear{samangouei2018defense}). % and compare the performance of using our triplet regularization or not. 
As shown in Table 3, when attacked by C\&W, model B attains 8.4\% error rate using Defense-GAN, while combining with triplet regularization, it achieves 4.7\% error rate. Again, this result shows that equipping the original defense method with triplet regularization can make the trained model more robust.

\section{Conclusion}
    In this paper, we propose Adversarial Training with Triplet Loss (AT$^2$L), which incorporates a modified triplet loss in the adversarial training process to alleviate the distortion of the models' classification boundary. We further design an ensemble version of AT$^2$L and propose to use the triple loss as a regularization term. The results of our experiments validate the effectiveness of our algorithms and demonstrate that our triplet regularization can be applied to existing defense methods for further improvement of robustness. 
    
\section*{Acknowledgments}
This work was partially supported by the National Key R\&D Program of China (2018YFB1004300), NSFC-NRF Joint Research Project (61861146001), YESS (2017QNRC001), and the Collaborative Innovation Center of Novel Software Technology and Industrialization.

\bibliographystyle{named}
\bibliography{attack}
\appendix
\section*{Appendix}

We show more details about our experiments in this appendix. Our experiments mainly contain two parts, the first one is to examine whether the new triplet loss function can improve the performance of adversarial training or not, and the performance when triplet loss is taken as a regularization. The second one is to apply our new triplet loss function to current defense methods. We do experiments over three datasets for the first part, Cats vs. Dogs, MNIST and CIFAR10. For the second part, we follow the setting of the original papers of these defense methods.
	Our new triplet loss is defined as follows,

    \begin{equation*}
    \begin{split}
    	\hat\ell(\mathbf{x},y) =& \frac{1}{(1+\lambda_1)k} \Bigg( \sum_{i\in k}\ell(\mathbf{x}_i,y)+\lambda_1\sum_{i\in k}\ell(\mathbf{x}_i^{adv},y) \Bigg) \\
        &+ \frac{\lambda_2}{k}\sum_{i}\max\Bigg\{\|f(\mathbf{x}_i^{adv})-f(\mathbf{x}_i)\| \\
        &-\|f(\mathbf{x}_i^{adv})-f(\mathbf{x}_i^n)\|+\alpha,0\Bigg\}
    \end{split}
    \end{equation*}
\section{Adversarial Training with Triplet Loss (AT$^2$L)}
In this section, we show results over three datasets, Cats vs. Dogs, MNIST and CIFAR10. The attack methods used in this part are FGSM, I-FGSM, LL, I-LL and C\&W. We set $\epsilon$ as $0.3$, which is the scale of FGSM, I-FGSM, LL and I-LL. For iteration algorithms like I-FGSM and I-LL, we set the number of iteration as 10 and $\epsilon = 0.03$ for each iteration. The max-iteration of C\&W is set to be 1000. The initial constant $c$ is set to be $1$e$-3$ and the largest value of $c$ to go up to before giving up is $2$. The rate at which we increase constant is $2$ and the learning rate of C\&W attack is $5$e$-3$. The performance of our algorithm is stable in a large range of the hyper-parameters. $\lambda_1$ and $\lambda_2$ are designed to adjust the weight of the original loss, adversarial loss and triplet loss. $\alpha$ is designed to tune the margin between different classes. In our experiments, with fixed $\lambda_2$, the robustness of the model is stable when $\lambda_1$ varies from 0.2 to 1.5. With fixed $\lambda_1$, the robustness is stable when $\lambda_2$ varies from 0.1 to 5. As for $\alpha$ in the triplet loss, the robustness is stable when varying from 0.1 to 2.0 with step-size 0.1.

\subsection{Cats vs. Dogs}

We first do experiments on Cats vs. Dogs dataset. The model structures we chose are VGG16, VGG19, inceptionv3, resnet50. Here we use FGSM, I-FGSM, LL, and I-LL to generate adversarial examples and notate these methods as gradient training methods and we select C\&W as the optimization attack method for the purpose of adversarial training. The margin $\alpha$ and $\lambda$ is set to be 0.5 and 0.3. We train an ensemble model which covers VGG16, VGG19, inceptionv3, resnet50 because each picture of this dataset is 224x224x3 and simple networks may not work well. The accuracy of the model over clean data is 91.2\%.

The known type of attack is proposed to train an ensemble model over model VGG16, VGG19, inceptionv3 and resnet50 when the original model $f(\cdot)$ is VGG16. We then attack the model with the adversarial examples generated against model VGG16. The unknown type of attack is designed to train the ensemble model over model VGG19, inceptionv3 and resnet50 when the original model $f(\cdot)$ is VGG19. We then attack the model with the adversarial examples generated against model VGG16 so that the attackers are not ware of the model's structure during the training process.

The accuracy of adversarial training with triplet loss against gradient-based adversarial examples is 4.7\%, which is better than normal adversarial training without triplet loss. Even after 10 iterations of adversarial training, the algorithm of adversarial training with triplet loss is still better. When trained with optimization-based algorithm like C\&W, the robustness against gradient-based adversarial examples is worse than when trained with gradient-based algorithm. However, the mixed version of both gradient-based and optimization-based algorithm works well regardless of the type of the attack.

The parameters in the triplet loss function are set to be $\lambda_1 = 0.3, \lambda_2=1, \alpha = 1.0$, and the results are shown in Figure 1, Figure 2, Table 1, and Table 2. 
\begin{table*}[!htbp]
\centering

\label{l0}
\scalebox{0.9}[0.9]{
\begin{tabular}{|c|c|c|c|c|c|}
\hline
Attack method         & Training method & Adv. Train(1)&Adv. Train(10)   & AT$^2$L(1)&AT$^2$L(10) \\ \hline
\multirow{3}{*}{FGSM} & Gradient-based        & 18.3 & 11.6    &  13.6 & \textbf{8.3}     \\ \cline{2-6} 
                      & Optimization-based              & 32.6 & 26.5    &  27.4 & 16.3    \\ \cline{2-6} 
                      & Mixed        & 20.8 & 19.6    &  13.7 & 11.3     \\ \hline
\multirow{3}{*}{C\&W}   & Gradient-based        & 93.5 & 96.2    &  95.2 & 93.1     \\ \cline{2-6} 
                      & Optimization-based             & 27.4 & 24.1    &  16.2 & 13.7    \\ \cline{2-6} 
                      & Mixed        & 25.3 & 22.2    &  17.3 & \textbf{12.0}     \\ \hline
\end{tabular}}
\caption{Error rate of known type of attack on Cats vs. Dogs. We train over model VGG16, VGG19, inceptionv3, resnet50, and we test the error rate over model VGG16. Baseline is traditional adversarial training. We use different algorithm to generate adversarial examples for training.}
\end{table*}
\begin{figure*}[!htbp] 
\centering
\subfigure[Attack by FGSM]{
\includegraphics[width=7cm]{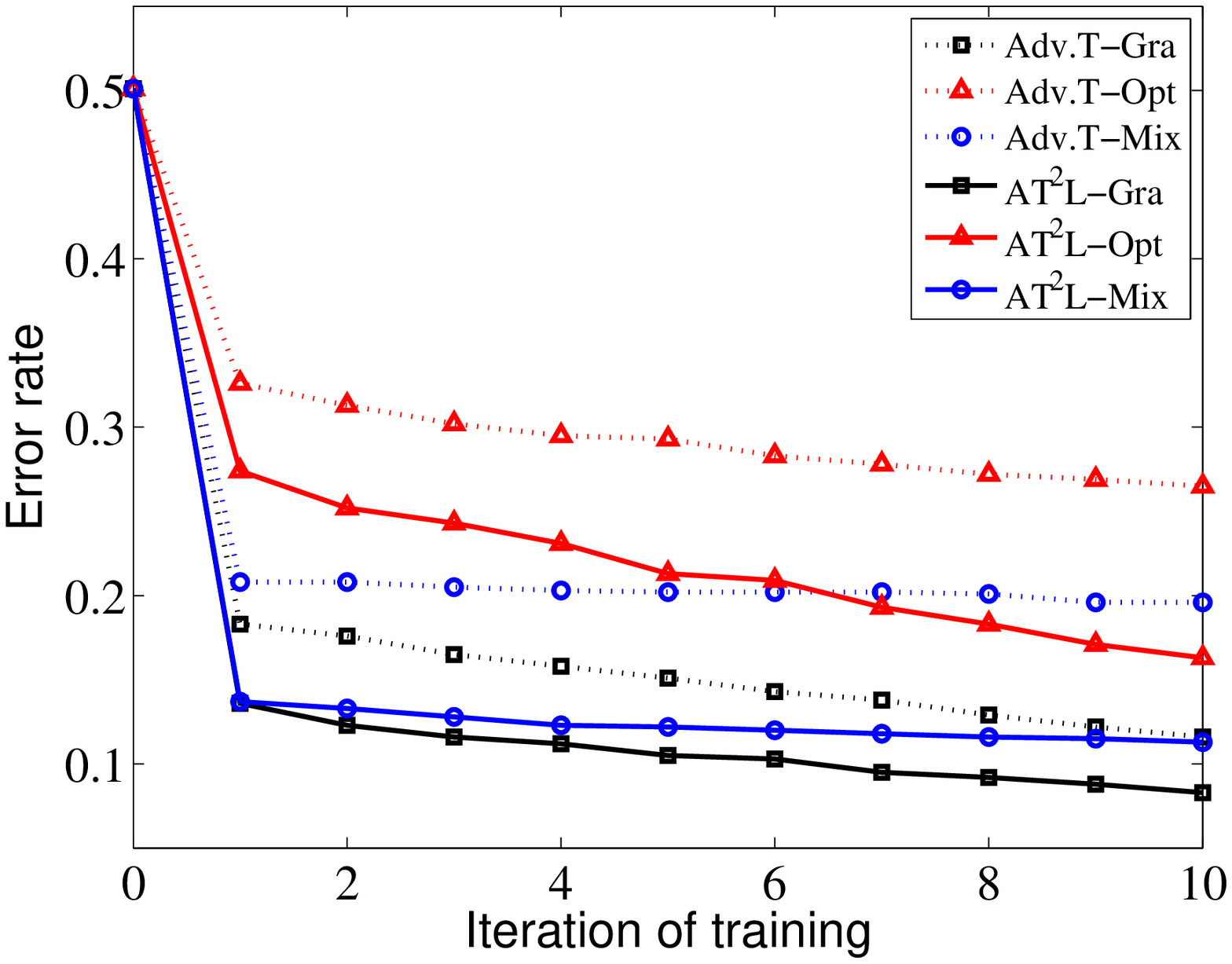}}
\subfigure[Attack by C\&W]{
\includegraphics[width=7cm]{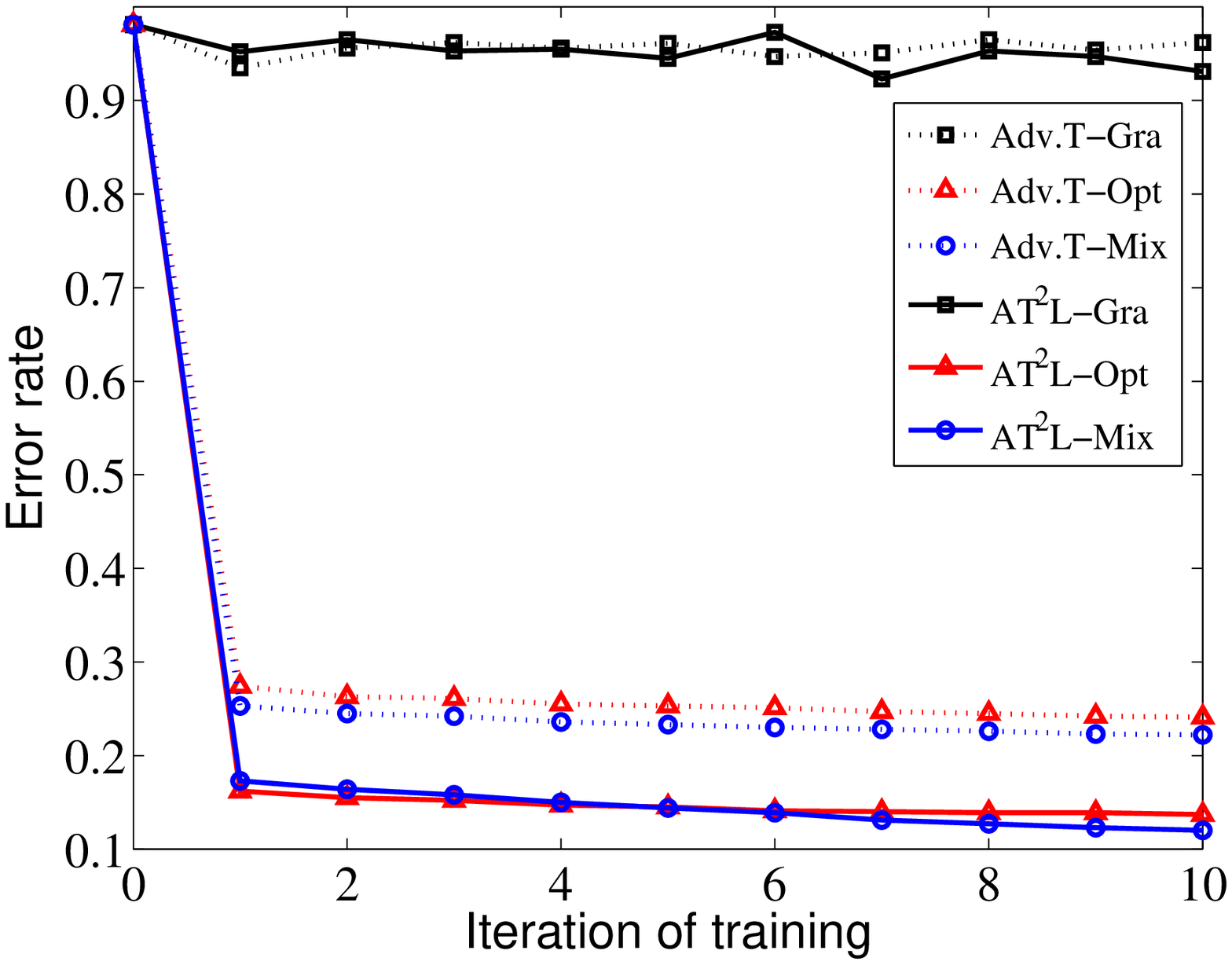}}

\caption{Results on Cats vs. Dogs dataset. `Adv.T' means traditional adversarial training. `-Gra' means the training process use the gradient-based algorithms to generate adversarial examples. `-Opt' means optimization-based algorithm C\&W and `-Mix' means mixed version of algorithm.}
\label{fig:2} 
\end{figure*}

\begin{table*}[!htbp]
\centering

\label{l01}
\scalebox{0.9}[0.9]{
\begin{tabular}{|c|c|c|c|c|c|}
\hline
Attack method         & Training method & Adv. Train(1)&Adv. Train(10)   & AT$^2$L(1)&AT$^2$L(10) \\ \hline
\multirow{3}{*}{FGSM} & Gradient-based        & 42.6 & 25.7    &  21.5 & 15.2     \\ \cline{2-6} 
                      & Optimization-based              & 43.7 & 35.9   &  32.9 & 19.2    \\ \cline{2-6} 
                      & Mixed        & 41.3 & 22.7    &  23.8 & \textbf{12.5}     \\ \hline
\multirow{3}{*}{C\&W}   & Gradient-based        & 99.3 & 95.4    &  92.1 & 94.6     \\ \cline{2-6} 
                      & Optimization-based              & 32.8 & 30.3    &  18.3 & 15.2    \\ \cline{2-6} 
                      & Mixed        & 27.6& 24.6    &  19.5 & \textbf{13.5}     \\ \hline
\end{tabular}}
\caption{Error rate of unknown type of attack on Cats vs. Dogs. We train over model VGG16, VGG19, inceptionv3, resnet50, and we test the error rate over model VGG16. Baseline is traditional adversarial training. We use different algorithm to generate adversarial examples for training.}
\end{table*}
\begin{figure*}[!htbp] 
\centering
\subfigure[Attack by FGSM]{
\includegraphics[width=7cm]{Cats_FGSM_B.eps}}
\subfigure[Attack by C\&W]{
\includegraphics[width=7cm]{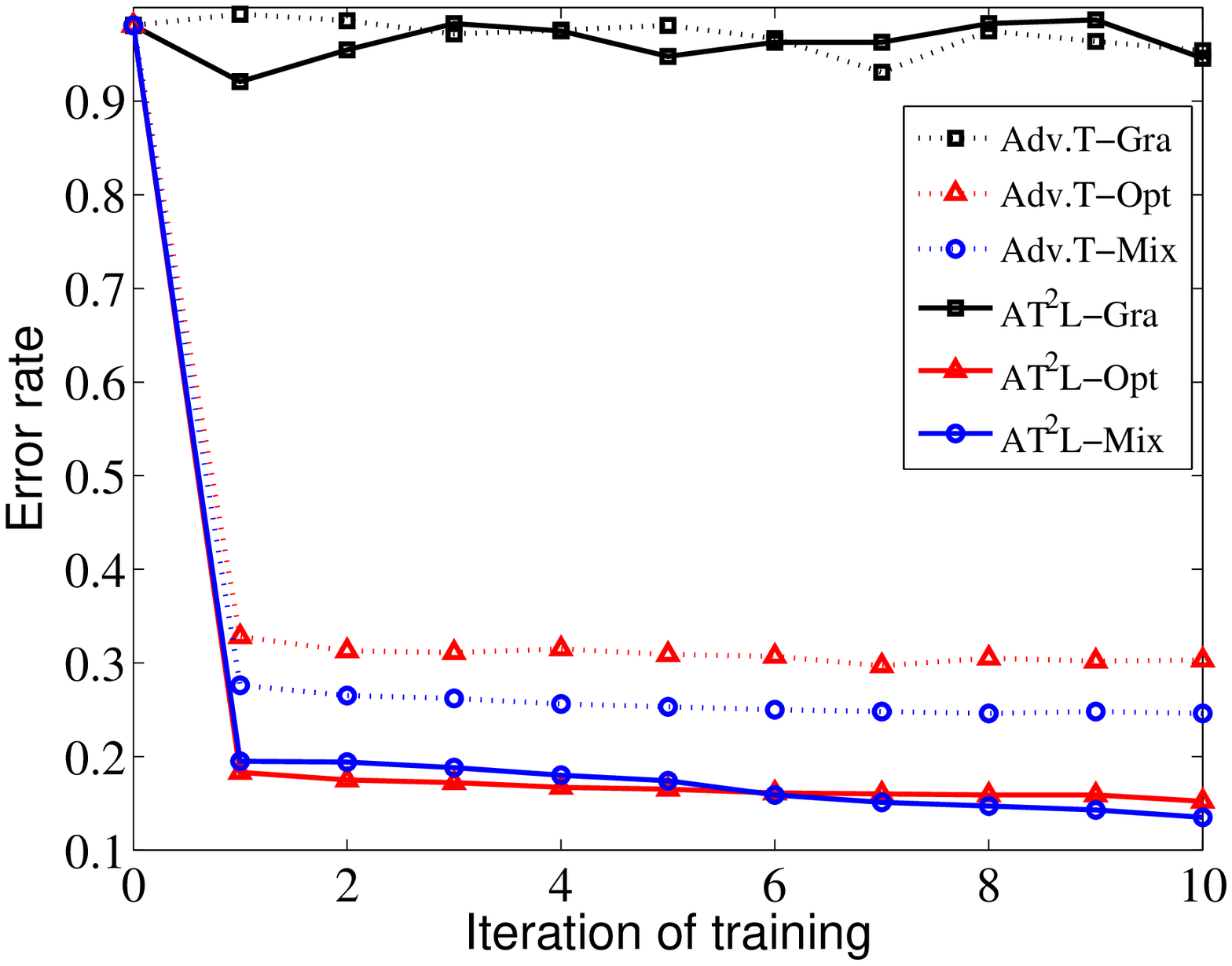}}

\caption{Results on Cats vs. Dogs dataset. `Adv.T' means traditional adversarial training. `-Gra' means the training process use the gradient-based algorithms to generate adversarial examples. `-Opt' means optimization-based algorithm C\&W and `-Mix' means mixed version of algorithm.}
\label{fig:2} 
\end{figure*}

\subsection{MNIST}

We then tested our algorithm on MNIST. The size of instance of MNIST is not as large as the instance of Cats vs. Dogs, but the number of classes is a bit larger. Each picture of MNIST is 28x28x1, so we construct 4 different models for the ensemble training process (which are shown in the Table 6). Model A, B, C are CNNs with different constructions, and model D only contains multiply dense layers and dropout layers. We perform a known type of attack and a unknown type of attack on this dataset.

The known type of attack is proposed to train an ensemble model over model A, B, C and D ($M$ in algorithm 2 is set to be [A, B, C, D]) and then attack the model with the adversarial examples generated against model A. Here we use FGSM, I-FGSM, LL, I-LL and a combination of these 4 algorithms to generate adversarial examples for training. The model we pre-trained (Step 1 of Algorithm 2) reaches up to 99.1\% accuracy. The parameter $k$ is 32, $\lambda_1$ is 0.3, $\lambda_2=1.0$ and $\alpha$ is set to be 1.0. The model after normal adversarial training or adversarial training with triplet loss does not lose the accuracy, its accuracy is still over 99\% on average. We also do more experiments about the performance of different gradient-based attack methods and their combination. The result shows that AT$^2$L can increase the robustness of the model. When training with the mixed version, the robustness against corresponding attack methods shows the best performance. 

The unknown type of attack is proposed to train an ensemble model over model A, B and D ($M$ in algorithm 2 is set to be [A, B, D]) and then attack the model with the adversarial examples generated against model C. Other settings are the same as the experiments of known type of attack. The result is quite similar as known type of attack, and the mixed version of AT$^2$L which is trained by adversarial examples against both gradient-based algorithms and optimization-based algorithms shows the lowest error rate.

The parameters in the triplet loss function are set to be $\lambda_1 = 0.3, \lambda_2=1, \alpha = 1.0$, and the results are shown in Figure 3, Figure 4, Table 4, and Table 5. The model structures we used in the experiments are shown in Table 3. 
\begin{table*}[!htbp]
\centering

\label{l1}
\begin{tabular}{|c|c|c|c|}
\hline
A         & B & C   & D \\ \hline
Input & Input            & Input    &  Input     \\ \hline
Conv(64,5,5) & Conv(64,8,8)            & Conv(128,3,3)    &  Dense(300)     \\ \hline
ReLu & ReLu            & ReLu    &  ReLu     \\ \hline
Conv(64,5,5) & Conv(128,6,6)            & Conv(64,3,3)    &  Dropout     \\ \hline
ReLu & ReLu            & ReLu    &  Dense(300)     \\ \hline
Dropout & Conv(128,5,5)            & Dropout    &  ReLu     \\ \hline
Dense(128) & ReLu            & Flatten    &  Dropout     \\ \hline
ReLu & Dropout            & Dense(128)    &  Dense(300)     \\ \hline
Dropout & Flatten            & ReLu    &  ReLu     \\ \hline
Dense(10) & Dense(10)            & Dropout    &  Dropout     \\ \hline
 &             & Dense(10)    &  Dense(10)     \\ \hline
\end{tabular}
\caption{Structures of models on MNIST.}
\end{table*}
\begin{table*}[!htbp]
\centering

\label{l2}
\scalebox{0.9}[0.9]{
\begin{tabular}{|c|c|c|c|c|c|}
\hline
Attack method         & Training method & Adv. Train(1)& Adv. Train(10)   & AT$^2$L(1)&AT$^2$L(10) \\ \hline
\multirow{6}{*}{FGSM} & FGSM            & 56.9 & 42.0    &  54.0 & \textbf{34.8 }    \\ \cline{2-6} 
                      & iter\_FGSM      & 83.3 & 88.3    &  66.6 & 38.3     \\ \cline{2-6} 
                      & LL              & 56.7 & 48.1    &  46.2 & 42.9     \\ \cline{2-6} 
                      & iter\_LL        & 87.9 & 83.8    &  53.0 & 44.1     \\ \cline{2-6}
                      & Gradient-based        & 58.3 & 46.3    &  51.8 & 40.3     \\ \cline{2-6} 
                      & Optimization-based              & 85.2 & 84.2    &  68.9 & 59.6     \\ \cline{2-6} 
                      & Mixed        & 86.7 & 54.9    &  55.9 & 37.2     \\ \hline
\multirow{6}{*}{C\&W}   & FGSM            & 91.6 & 86.8    &  91.1 & 88.6     \\ \cline{2-6} 
                      & iter\_FGSM      & 83.0 & 88.9    &  77.9 & 85.2     \\ \cline{2-6} 
                      & LL              & 82.6 & 91.3    &  86.8 & 88.5     \\ \cline{2-6} 
                      & iter\_LL        & 79.8 & 84.2    &  89.9 & 85.2     \\ \cline{2-6}
                      & Gradient-based        & 84.1 & 88.7    &  83.5 & 89.3     \\ \cline{2-6} 
                      & Optimization-based             & 56.4 & 56.1    &  43.3 & 45.2     \\ \cline{2-6} 
                      & Mixed        & 62.2 & 55.7    &  39.5 & \textbf{23.5}     \\ \hline
\end{tabular}}
\caption{Error rate of known type of attack on MNIST. We train over model A,B,C,D, and we test the error rate over model A. Baseline is traditional adversarial training. We use different algorithm to generate adversarial examples for training. We then use FGSM or C\&W for testing the robustness.}
\end{table*}
\begin{figure*}[!htbp] 
\centering
\subfigure[Attack by FGSM]{
\includegraphics[width=7cm]{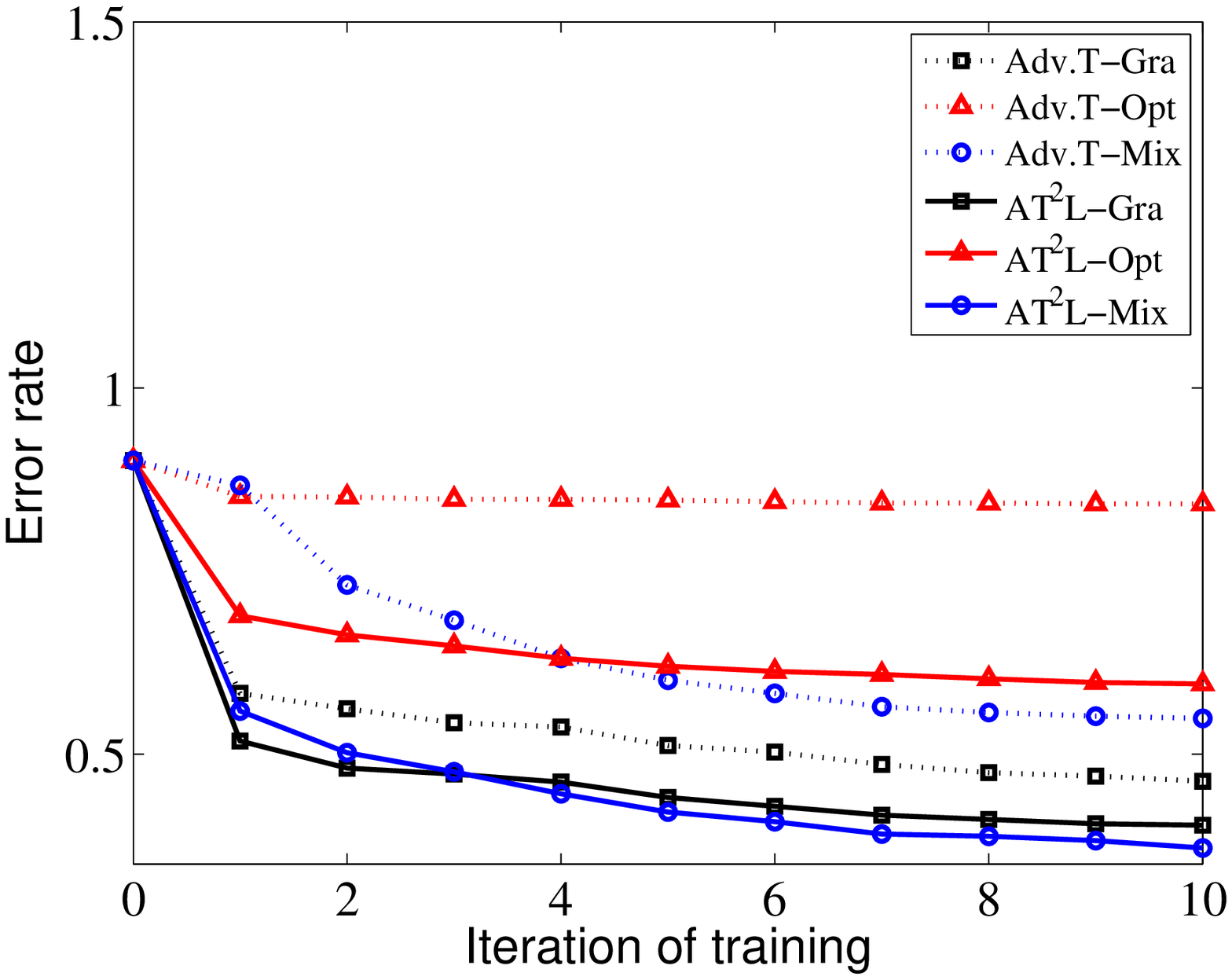}}
\subfigure[Attack by C\&W]{
\includegraphics[width=7cm]{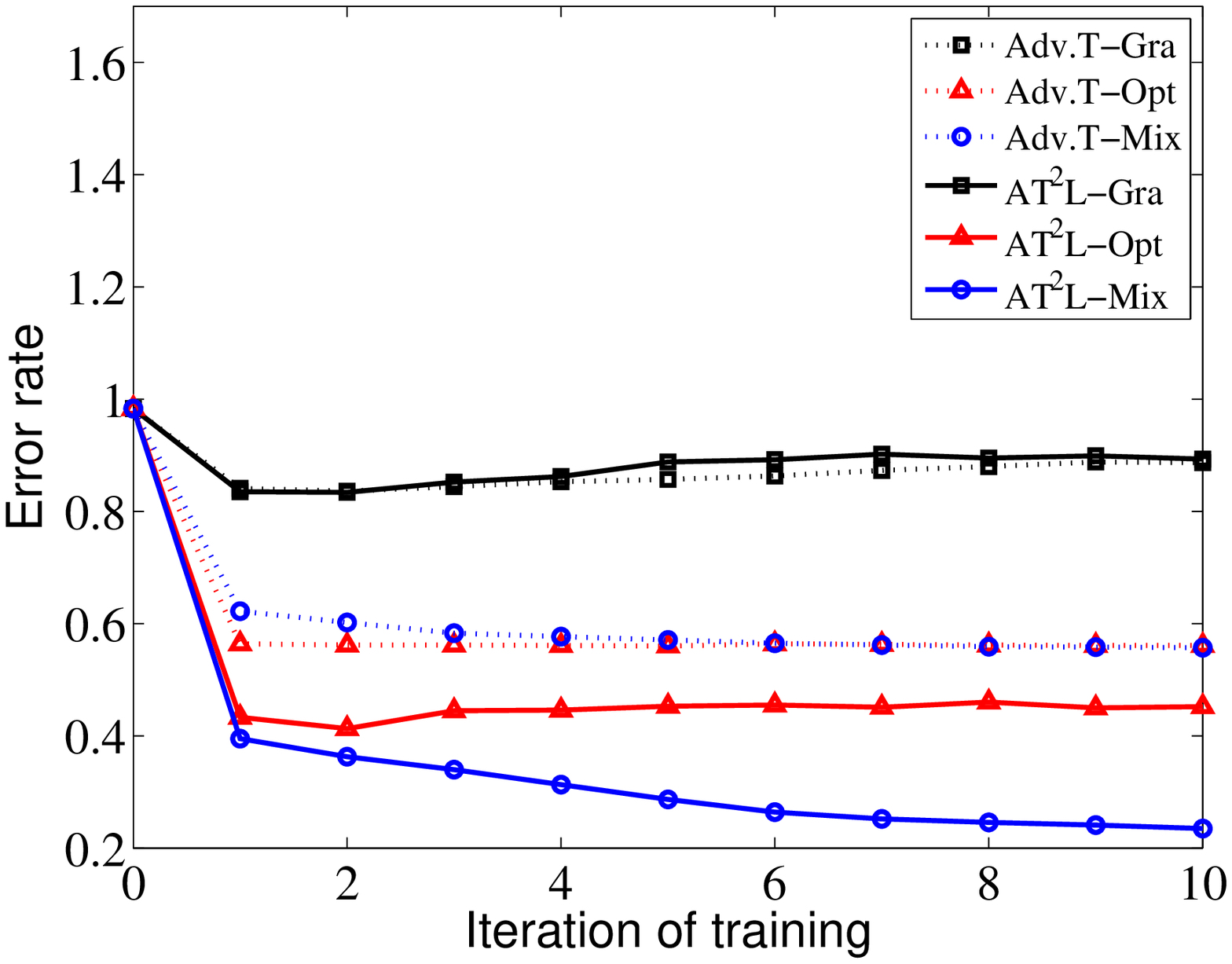}}

\caption{Results on MNIST dataset of known type of attack. `Adv.T' means traditional adversarial training. `-Gra' means the training process use the gradient-based algorithms to generate adversarial examples. `-Opt' means optimization-based algorithm C\&W and `-Mix' means mixed version of algorithm.}
\label{fig:3} 
\end{figure*} 

\begin{table*}[!htbp]
\centering

\label{l3}
\scalebox{0.9}[0.9]{
\begin{tabular}{|c|c|c|c|c|c|}
\hline
Attack method         & Training method & Adv. Train(1)& Adv. Train(10)   & AT$^2$L(1)&AT$^2$L(10) \\ \hline
\multirow{6}{*}{FGSM} & FGSM            & 59.3 & 52.6    &  52.1 & 45.2     \\ \cline{2-6} 
                      & iter\_FGSM      & 83.5 & 85.8    &  63.6 & 49.1     \\ \cline{2-6} 
                      & LL              & 62.1 & 54.7    &  45.3 & \textbf{38.5}     \\ \cline{2-6} 
                      & iter\_LL        & 82.5 & 86.1    &  51.9 & 45.1     \\ \cline{2-6} 
                      & Gradient-based        & 67.4 & 53.5    &  54.1 & 42.2     \\ \cline{2-6}                       & Optimization-based               & 86.1 & 87.8    &  59.2 & 55.3     \\ \cline{2-6} 
                      & Mixed        & 61.4 & 58.2    &  51.6 & 39.1     \\ \hline
\multirow{6}{*}{C\&W}   & FGSM            & 95.1 & 97.5    &  95.2 & 92.9     \\ \cline{2-6} 
                      & iter\_FGSM      & 95.8 & 94.2    &  87.2 & 89.1     \\ \cline{2-6} 
                      & LL              & 94.1 & 93.2    &  93.9 & 86.6     \\ \cline{2-6} 
                      & iter\_LL        & 97.9 & 94.7    &  98.5 & 83.1     \\ \cline{2-6} 
                      & Gradient-based        & 91.3 & 95.1    &  95.9 & 97.2     \\ \cline{2-6}                       & Optimization-based               & 58.3 & 57.8    &  46.1 & 43.3     \\ \cline{2-6} 
                      & Mixed        & 64.9 & 53.1    &  47.3 & \textbf{32.7}     \\ \hline
\end{tabular}}
\caption{Error rate of unknown type of attack on MNIST. We train over model A,B,D, and we test the error rate over model C. Baseline is traditional adversarial training. We use different algorithm to generate adversarial examples for training. We then use FGSM or C\&W for testing the robustness. }
\end{table*}
\begin{figure*}[!htbp] 
\centering
\subfigure[Attack by FGSM]{
\includegraphics[width=7cm]{MNIST_FGSM_BLACK.eps}}
\subfigure[Attack by C\&W]{
\includegraphics[width=7cm]{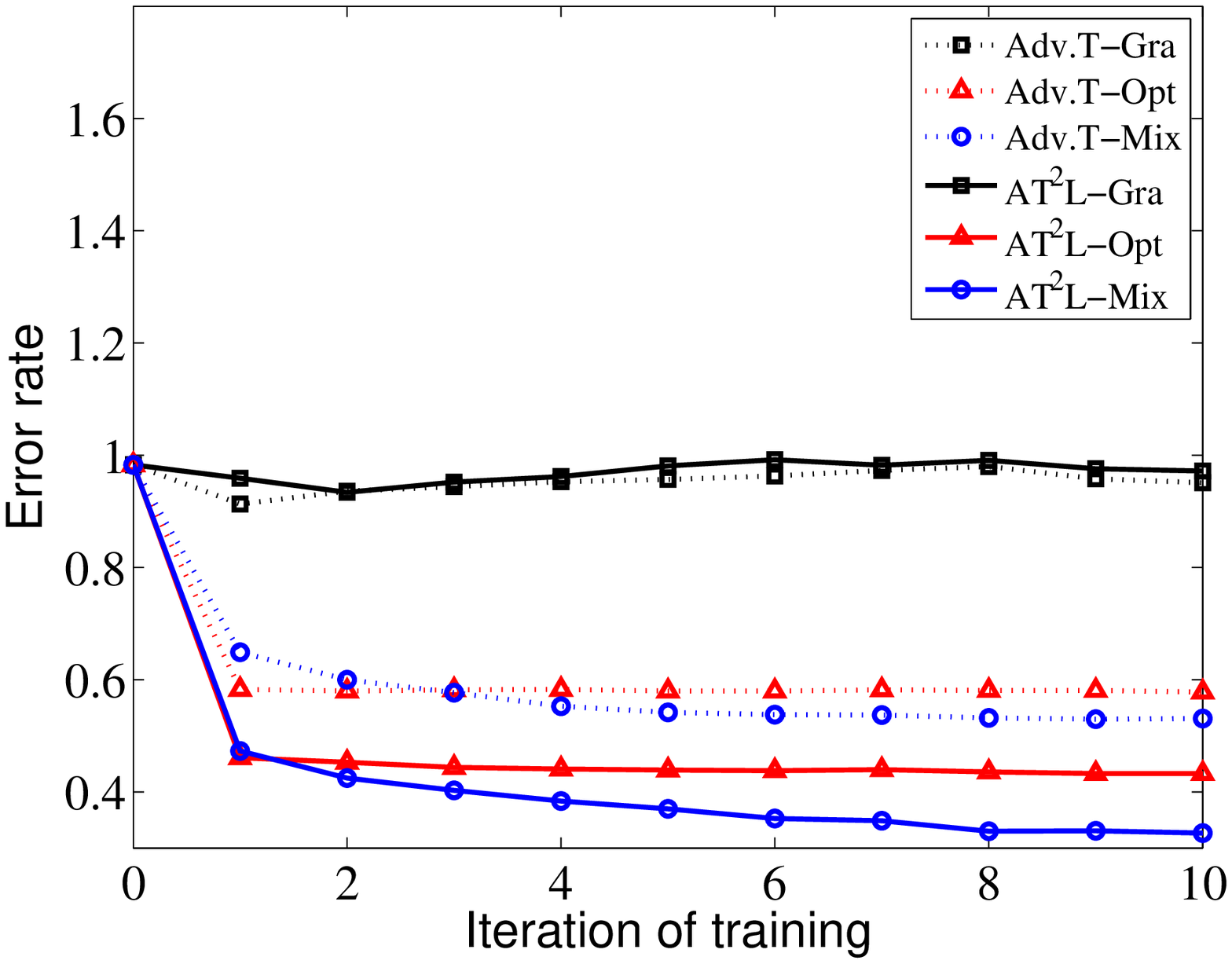}}

\caption{Results on MNIST dataset of unknown type of attack. `Adv.T' means traditional adversarial training. `-Gra' means the training process use the gradient-based algorithms to generate adversarial examples. `-Opt' means optimization-based algorithm C\&W and `-Mix' means mixed version of algorithm.}
\label{fig:4} 
\end{figure*} 

\subsection{CIFAR10}
MNIST is a standard dataset in the research of adversarial examples, but it's a small dataset and we further do more experiments over CIFAR10. The models we choose for CIFAR10 are the same as the Cats vs. Dogs dataset: VGG16, VGG19, inceptionv3 and resnet50. 

We also do a known type of attack and a unknown type of attack on CIFAR10. The known type of attack proposes to train an ensemble model over model VGG16, VGG19, inceptionv3 and resnet50 ($M$ in algorithm 2 is set to be [VGG16, VGG19, inceptionv3, resnet50]) and then attack the model with the adversarial examples generated against model VGG16.  The unknown type of attack is proposed to train over models mentioned above without VGG16 and attack the model against VGG16. The setting of parameters over both attacks are the same: $k=32,\alpha = 1.0, \lambda_1 = 0.6, \lambda_2=1.0$. The accuracy of the pre-trained model (Step 1 of Algorithm 2) reaches up to 92.6\% and the accuracy does not decrease much after adversarial training (near 92\%). The results of unknown type of attack are a little worse than the known type of attack, but the robustness of the model indeed increases. The error rate of the original model against FGSM is 90.1\% and that of the original model against C\&W is 99.2\%. From the results we can see that our AT$^2$L trained over ensemble models and mixed attack methods shows the best performance against both gradient-based and optimization-based attack. The triplet loss can increase the robustness of the model better than the normal adversarial training without triplet loss.

The model structures used to generate adversarial examples are the same as the setting of dataset Cats vs. Dogs. We experiment both known type of and unknown type of settings and the result shows that our special triplet loss can improve the robustness of the model better than traditional adversarial training process. The results are shown in Figure 5, Figure 6, Table 6, and Table 7. 
\begin{table*}[!htbp]
\centering

\label{l4}
\scalebox{0.9}[0.9]{
\begin{tabular}{|c|c|c|c|c|c|}
\hline
Attack method         & Training method & Adv. Train(1)& Adv. Train(10)   & AT$^2$L(1)&AT$^2$L(10) \\ \hline
\multirow{3}{*}{FGSM} & Gradient-based        & 64.2 & 50.3    &  49.6 & \textbf{41.8}     \\ \cline{2-6} 
                      & Optimization-based               & 74.3 & 81.9    &  63.1 & 62.8     \\ \cline{2-6} 
                      & Mixed        & 66.3 & 63.7    &  57.2 & 44.3     \\ \hline
                      
\multirow{3}{*}{C\&W}   & Gradient-based        & 97.1 & 98.8    &  96.3 & 97.9     \\ \cline{2-6} 
                      & Optimization-based               & 53.9 & 43.0    &  46.1 & \textbf{35.7}     \\ \cline{2-6} 
                      & Mixed        & 62.5 & 61.1    &  53.9 & 44.4     \\ \hline
\end{tabular}}
\caption{Error rate of known type of attack on CIFAR10. We train over model VGG16, VGG19, inceptionv3, resnet50, and we test the error rate over model VGG16. Baseline is traditional adversarial training. We use different algorithm to generate adversarial examples for training. We then use FGSM or C\&W for testing the robustness. }
\end{table*}
\begin{figure*}[!htbp] 
\centering
\subfigure[Attack by FGSM]{
\includegraphics[width=7cm]{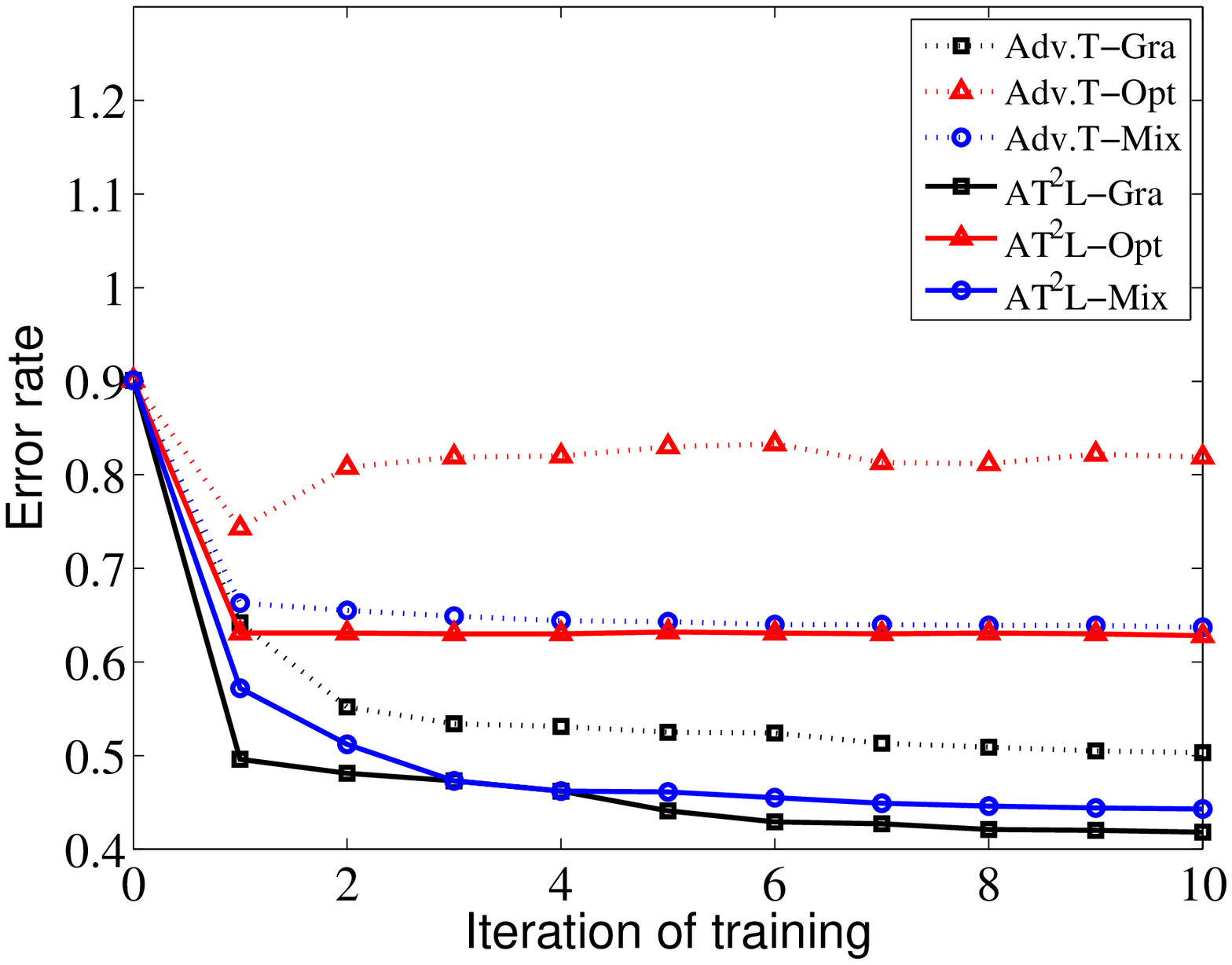}}
\subfigure[Attack by C\&W]{
\includegraphics[width=7cm]{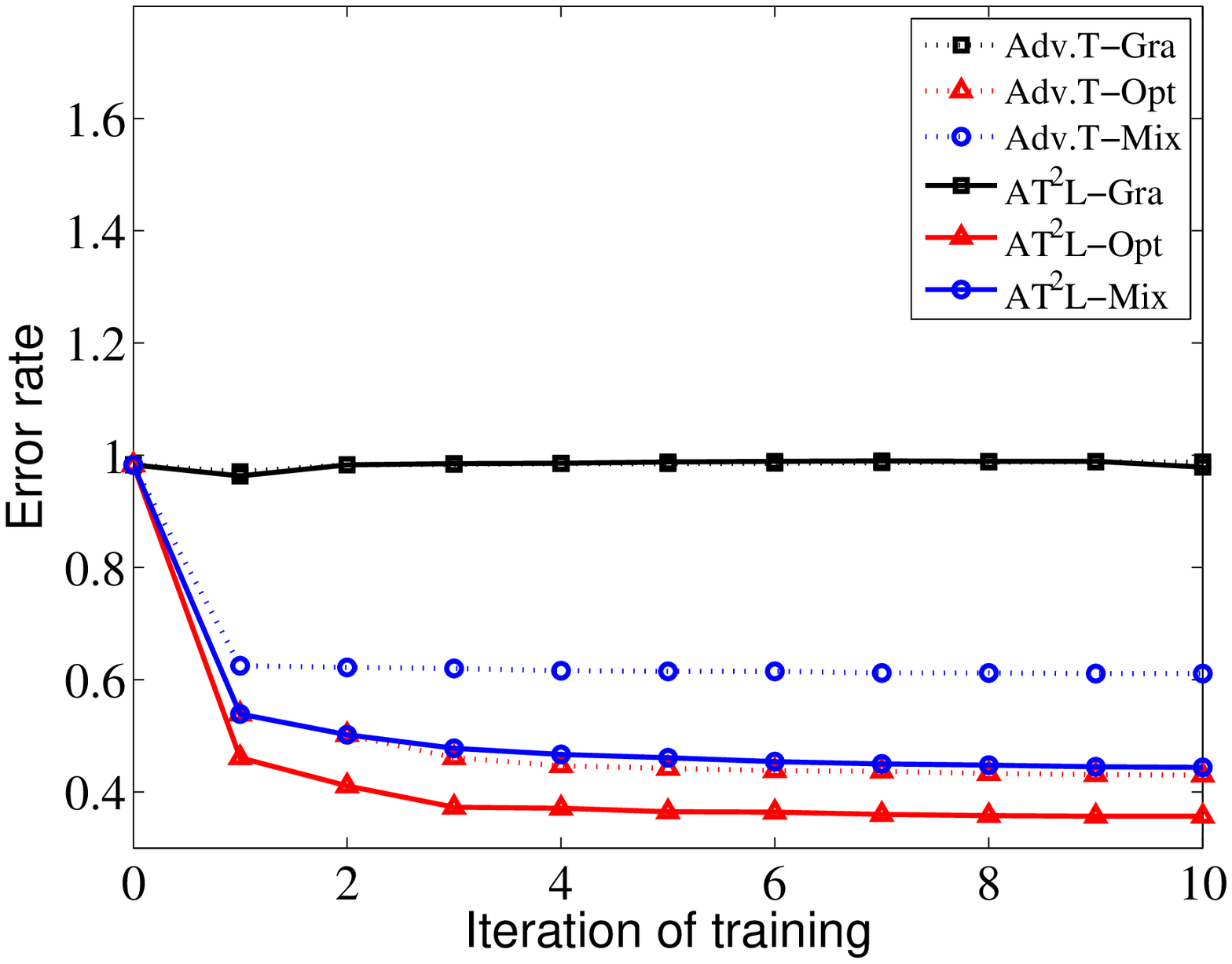}}

\caption{Results on CIFAR10 dataset of known type of attack. `Adv.T' means traditional adversarial training. `-Gra' means the training process use the gradient-based algorithms to generate adversarial examples. `-Opt' means optimization-based algorithm C\&W and `-Mix' means mixed version of algorithm.}
\label{fig:5} 
\end{figure*}

\begin{table*}[!htbp]
\centering

\label{l5}
\scalebox{0.9}[0.9]{
\begin{tabular}{|c|c|c|c|c|c|}
\hline
Attack method         & Training method & Adv. Train(1)&Adv. Train(10)   & AT$^2$L(1)&AT$^2$L(10) \\ \hline
\multirow{3}{*}{FGSM} & Gradient-based        & 67.6 & 53.1    &  45.1 & \textbf{41.6}     \\ \cline{2-6} 
                      & Optimization-based              & 84.9 & 88.3    &  77.3 & 71.4     \\ \cline{2-6} 
                      & Mixed        & 73.8 & 69.2    &  61.7 & 50.9     \\ \hline
                      
\multirow{3}{*}{C\&W}   & Gradient-based        & 98.5 & 97.3    &  98.5 & 98.8     \\ \cline{2-6} 
                      & Optimization-based              & 57.6 & 50.7    &  51.1 & \textbf{45.7}     \\ \cline{2-6} 
                      & Mixed        & 69.4 & 63.6    &  56.8 & 47.3     \\ \hline
\end{tabular}}
\caption{Error rate of unknown type of attack on CIFAR10. We train over model VGG19, inceptionv3, resnet50, and we test the error rate over model VGG16. Baseline is traditional adversarial training. We use different algorithm to generate adversarial examples for training. We then use FGSM or C\&W for testing the robustness. }
\end{table*}

\begin{figure*}[!htbp] 
\centering
\subfigure[Attack by FGSM]{
\includegraphics[width=7cm]{CIFAR_FGSM_BLACK.eps}}
\subfigure[Attack by C\&W]{
\includegraphics[width=7cm]{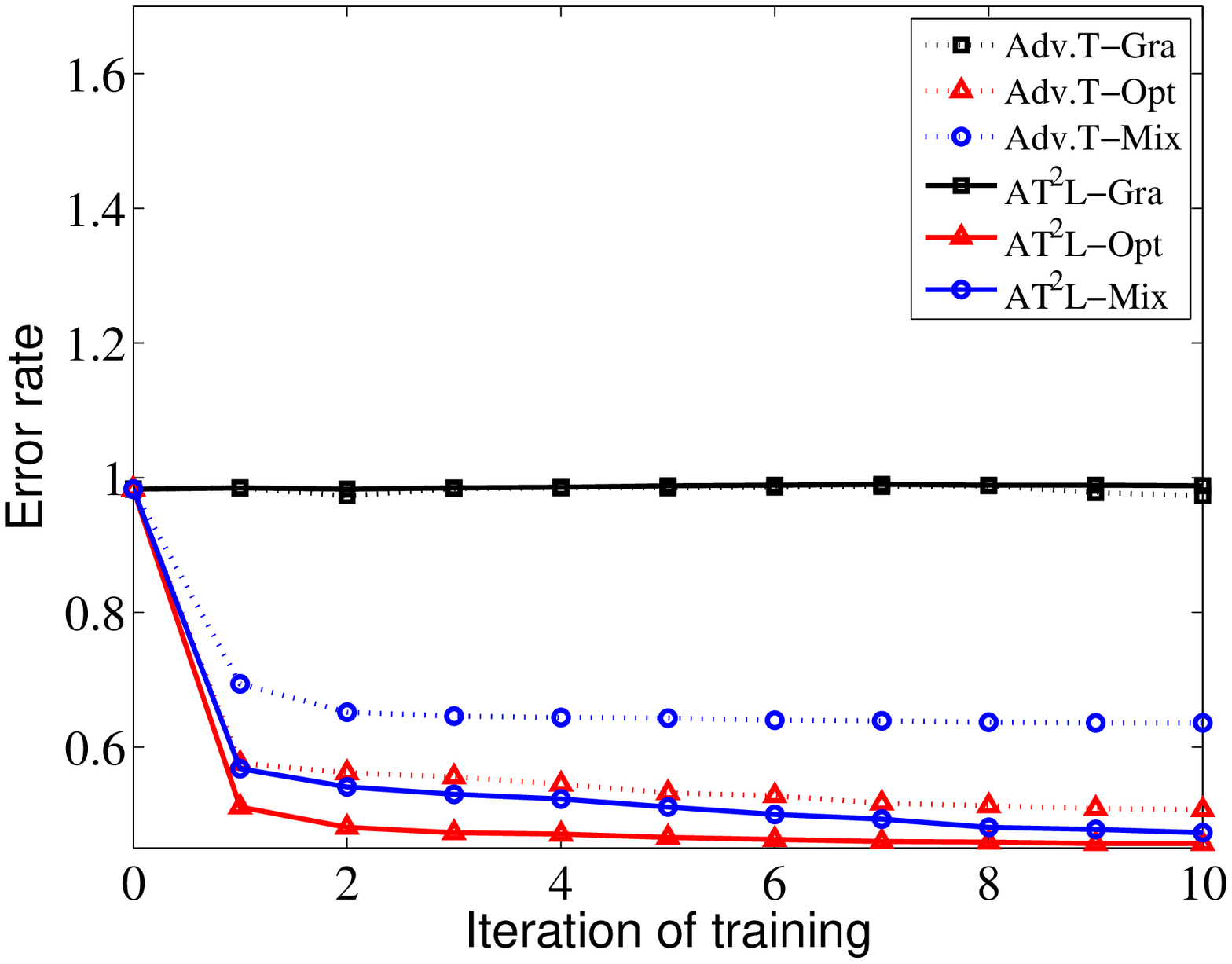}}

\caption{Results on CIFAR10 dataset of unknown type of attack. `Adv.T' means traditional adversarial training. `-Gra' means the training process use the gradient-based algorithms to generate adversarial examples. `-Opt' means optimization-based algorithm C\&W and `-Mix' means mixed version of algorithm.}
\label{fig:6} 
\end{figure*} 

\section{Apply to current defense}
The second part of our experiment is to apply our new loss to the existence defense methods. We choose three typical defense methods described in the  paper, thermometer encoding,  mitigating through randomization, and Defense-GAN. We propose to improve the robustness of the model based on these defenses. %The method used for generated adversarial examples for triplet regularization is C\&W.

\subsection{Thermometer encoding}
Thermometer encoding used a phenomenon called Gradient Shattering. This defense is designed to encode the original input to a non-differentiable image and the original defense is combined with a traditional adversarial training process. So our triplet regularization can be easily inserted into this defense. We only need to change the loss function during the adversarial training process to our new loss function and we do not need to generate different types of adversarial examples or against different model structures. In this experiment, we view the image after the thermometer encoding as the adversarial examples. In the known type of setting, we use the same model structure in both training and testing process and in the unknown type of setting we apply different structures. We show the results of the first round of adversarial training and the results after 7 rounds of training.

The parameters in the triplet loss function are set to be $\lambda_1 = 0.3, \lambda_2=0.7, \alpha = 1.5$, and the results are shown in Table 8-11. In each table, we test our model with three types of examples, i.e. clean data, examples generated by FGSM, and examples generated by PGD\/LS-PGA. 
\begin{table*}[!htbp]
\centering
\label{l8}
\scalebox{0.9}[0.9]{
\begin{tabular}{|c|c|c|c|c|c|c|c|c|c|c|c|c|c|c|c|}
\hline
      & \multicolumn{3}{c|}{Ori}   & \multicolumn{3}{c|}{Th. En.(1)} & \multicolumn{3}{c|}{Th. En.(7)} & \multicolumn{3}{c|}{Th. En.+ Tri. Reg(1)} & \multicolumn{3}{c|}{Th. En.+ Tri. Reg(7)} \\ \hline
Clean & \multicolumn{3}{c|}{\textbf{0.8}}   & \multicolumn{3}{c|}{\textbf{0.8}}           & \multicolumn{3}{c|}{0.9}           & \multicolumn{3}{c|}{\textbf{0.8}}                     & \multicolumn{3}{c|}{\textbf{0.8}}                     \\ \hline
FGSM  & \multicolumn{3}{c|}{100.0} & \multicolumn{3}{c|}{31.6}          & \multicolumn{3}{c|}{4.2}           & \multicolumn{3}{c|}{28.7}                    & \multicolumn{3}{c|}{\textbf{3.5}}                     \\ \hline
PGD/LS-PGA  & \multicolumn{3}{c|}{100.0} & \multicolumn{3}{c|}{25.0}          & \multicolumn{3}{c|}{5.9}           & \multicolumn{3}{c|}{23.0}                    & \multicolumn{3}{c|}{\textbf{2.7}}                     \\ \hline
\end{tabular}}
\caption{Error rate of known type of attacks on MNIST over Thermometer Models. `Th. En.' mean Thermometer Encoding. `Tri. Reg' means applying our triplet regularization.}
\end{table*}

\begin{table*}[!htbp]
\centering
\label{l9}
\scalebox{0.9}[0.9]{
\begin{tabular}{|c|c|c|c|c|c|c|c|c|c|c|c|c|c|c|c|}
\hline
      & \multicolumn{3}{c|}{Ori}   & \multicolumn{3}{c|}{Th. En.(1)} & \multicolumn{3}{c|}{Th. En.(7)} & \multicolumn{3}{c|}{Th. En.+ Tri. Reg(1)} & \multicolumn{3}{c|}{Th. En.+ Tri. Reg(7)} \\ \hline
Clean & \multicolumn{3}{c|}{\textbf{5.8}}  & \multicolumn{3}{c|}{7.6}           & \multicolumn{3}{c|}{10.1}          & \multicolumn{3}{c|}{6.1}                     & \multicolumn{3}{c|}{7.2}                     \\ \hline
FGSM  & \multicolumn{3}{c|}{51.5} & \multicolumn{3}{c|}{37.1}          & \multicolumn{3}{c|}{20.0}          & \multicolumn{3}{c|}{27.6}                    & \multicolumn{3}{c|}{\textbf{15.1}}                    \\ \hline
PGD/LS-PGA  & \multicolumn{3}{c|}{49.5} & \multicolumn{3}{c|}{39.3}          & \multicolumn{3}{c|}{20.9}          & \multicolumn{3}{c|}{29.7}                    & \multicolumn{3}{c|}{\textbf{13.4}}                    \\ \hline
\end{tabular}}
\caption{Error rate of known type of attacks on CIFAR10 over Thermometer Models. `Th. En.' mean Thermometer Encoding. `Tri. Reg' means applying our triplet regularization.}
\end{table*}

\begin{table*}[!htbp]
\centering
\label{l10}
\scalebox{0.9}[0.9]{
\begin{tabular}{|c|c|c|c|c|c|c|c|c|c|c|c|c|c|c|c|}
\hline
      & \multicolumn{3}{c|}{Ori}   & \multicolumn{3}{c|}{Th. En.(1)} & \multicolumn{3}{c|}{Th. En.(7)} & \multicolumn{3}{c|}{Th. En.+ Tri. Reg(1)} & \multicolumn{3}{c|}{Th. En.+ Tri. Reg(7)} \\ \hline

Clean      & \multicolumn{3}{c|}{\textbf{3.5}}  & \multicolumn{3}{c|}{5.1}           & \multicolumn{3}{c|}{5.5}           & \multicolumn{3}{c|}{5.0}                     & \multicolumn{3}{c|}{9.2}                     \\ \hline
FGSM       & \multicolumn{3}{c|}{89.1} & \multicolumn{3}{c|}{73.7}          & \multicolumn{3}{c|}{67.0}          & \multicolumn{3}{c|}{53.1}                    & \multicolumn{3}{c|}{\textbf{47.9}}                    \\ \hline
PGD/LS-PGA & \multicolumn{3}{c|}{88.2} & \multicolumn{3}{c|}{77.1}          & \multicolumn{3}{c|}{71.9}          & \multicolumn{3}{c|}{62.8}                    & \multicolumn{3}{c|}{\textbf{55.3}}                    \\ \hline
\end{tabular}}
\caption{Error rate of unknown type of attacks on MNIST over Thermometer Models. `Th. En.' mean Thermometer Encoding. `Tri. Reg' means applying our triplet regularization.}
\end{table*}

\begin{table*}[!htbp]
\centering
\label{l11}
\scalebox{0.9}[0.9]{
\begin{tabular}{|c|c|c|c|c|c|c|c|c|c|c|c|c|c|c|c|}
\hline
      & \multicolumn{3}{c|}{Ori}   & \multicolumn{3}{c|}{Th. En.(1)} & \multicolumn{3}{c|}{Th. En.(7)} & \multicolumn{3}{c|}{Th. En.+ Tri. Reg(1)} & \multicolumn{3}{c|}{Th. En.+ Tri. Reg(7)} \\ \hline

Clean      & \multicolumn{3}{c|}{\textbf{11.5}} & \multicolumn{3}{c|}{13.6}          & \multicolumn{3}{c|}{20.1}          & \multicolumn{3}{c|}{16.2}                    & \multicolumn{3}{c|}{17.7}                    \\ \hline
FGSM       & \multicolumn{3}{c|}{46.5} & \multicolumn{3}{c|}{43.8}          & \multicolumn{3}{c|}{39.1}          & \multicolumn{3}{c|}{38.6}                    & \multicolumn{3}{c|}{\textbf{33.0}}                    \\ \hline
PGD/LS-PGA & \multicolumn{3}{c|}{55.0} & \multicolumn{3}{c|}{52.9}          & \multicolumn{3}{c|}{49.3}          & \multicolumn{3}{c|}{51.9}                    & \multicolumn{3}{c|}{\textbf{43.0}}                    \\ \hline
\end{tabular}}
\caption{Error rate of unknown type of attacks on CIFAR10 over Thermometer Models. `Th. En.' mean Thermometer Encoding. `Tri. Reg' means applying our triplet regularization.}
\end{table*}

\subsection{Mitigating through randomization}
Mitigating through randomization uses Stochastic Gradients which are caused by randomized defenses. The input is randomly transformed before being fed to the classifier, causing the gradients to become randomized.  The target models and the defense models are exactly the same except for the parameter settings of the randomization layers, i.e., the randomization parameters of the target models are predefined while randomization parameters of the defense models are randomly generated at test time. 

The traditional adversarial training process is also mentioned in the original paper. So our special triplet regularization can be easily applied to this defense. We also drop the use of different types of  adversarial examples and different model structures used to improve the effect of adversarial training. We select two attack scenarios, the vanilla attack scenario and the ensemble-pattern attack scenario in our experiment.

The parameters in the triplet loss function are set to be $\lambda_1 = 0.3, \lambda_2=1, \alpha = 2.0$, and the results are shown in Table 12-13. We use three different attack methods, i.e., FGSM, Deepfool, and C\&W, to test the effect of the defense.

\begin{table*}[!htbp]
\centering

\label{l6}
\scalebox{0.75}[0.8]{
\begin{tabular}{|c|c|c|c|c|c|c|c|c|c|c|c|c|}
\hline
Models   & \multicolumn{3}{c|}{Inception-v3}                                  & \multicolumn{3}{c|}{ResNet-v2-101}                                 & \multicolumn{3}{c|}{Inception-ResNet-v2}                            & \multicolumn{3}{c|}{Ens-adv-Inception-ResNet-v2}                    \\ \hline
         & Ori   & Rand & \begin{tabular}[c]{@{}l@{}}Rand+\\Tri. Reg\end{tabular} & Ori   & Rand & \begin{tabular}[c]{@{}l@{}}Rand+\\Tri. Reg\end{tabular} & Ori   & Rand & \begin{tabular}[c]{@{}l@{}}Rand+\\ Tri. Reg\end{tabular} & Ori  & Rand & \begin{tabular}[c]{@{}l@{}}Rand+\\Tri. Reg\end{tabular} \\ \hline
FGSM     & 66.8  & 36.2 & \textbf{30.5}                                                & 73.7  & 28.2 & \textbf{21.7}                                                & 34.7  & 19.0 & \textbf{8.3}                                                 & 15.6 & \textbf{4.3}  & 4.6                                                 \\ \hline
Deepfool & 100.0 & 1.7  & \textbf{1.1}                                                 & 100.0 & 2.3  & \textbf{1.5}                                                 & 100.0 & 1.8  & \textbf{0.8}                                                 & 99.8 & 0.9  & \textbf{0.7}                                                 \\ \hline
C\&W     & 100.0 & 3.1  & \textbf{2.6}                                                 & 100.0 & 2.9  & \textbf{1.2}                                                 & 99.7  & 2.3  & \textbf{1.3}                                                 & 99.1 & 1.2  & \textbf{0.9}                                                 \\ \hline
\end{tabular}}
\caption{Top-1 classification error rate under the vanilla attack scenario. `Ori' means the original model. `Rand' means adding some randomization layers. `Tri. Reg' means applying our triplet regularization.}
\end{table*}

\begin{table*}[!htbp]
\centering

\label{l7}
\scalebox{0.77}[0.8]{
\begin{tabular}{|c|c|c|c|c|c|c|c|c|c|c|c|c|}
\hline
Models   & \multicolumn{3}{c|}{Inception-v3}                                 & \multicolumn{3}{c|}{ResNet-v2-101}                                & \multicolumn{3}{c|}{Inception-ResNet-v2}                           & \multicolumn{3}{c|}{Ens-adv-Inception-ResNet-v2}                    \\ \hline
         & Ori  & Rand & \begin{tabular}[c]{@{}l@{}}Rand+\\Tri. Reg\end{tabular} & Ori  & Rand & \begin{tabular}[c]{@{}l@{}}Rand+\\Tri. Reg\end{tabular} & Ori  & Rand & \begin{tabular}[c]{@{}l@{}}Rand+\\ Tri. Reg\end{tabular} & Ori  & Rand & \begin{tabular}[c]{@{}l@{}}Rand +\\Tri. Reg\end{tabular} \\ \hline
FGSM     & 62.7 & 58.8 & \textbf{37.3}                                                & 60.8 & 55.1 & \textbf{45.1}                                                & 28.5 & 25.7 & \textbf{23.5}                                                & 13.8 & 11.1 & \textbf{8.9}                                                 \\ \hline
Deepfool & 99.4 & 18.7 & \textbf{17.7}                                                & 99.1 & \textbf{19.5} & 22.2                                                & 99.1 &\textbf{ 30.6} & 41.6                                                & 98.4 & \textbf{6.5}  & 9.0                                                 \\ \hline
C\&W     & 99.4 & 37.1 & \textbf{23.1}                                                & 99.0 & 25.7 & \textbf{20.0}                                                & 98.4 & 31.7 & \textbf{21.7}                                                & 94.2 & 13.9 & \textbf{7.4}                                                 \\ \hline
\end{tabular}}
\caption{ Top-1 classification error rate under the ensemble-pattern attack scenario. Similar to vanilla
attack and single-pattern attack scenarios, we see that randomization layers increase the accuracy
under all attacks and networks. This clearly demonstrates the effectiveness of the proposed randomization
method on defending against adversarial examples, even under this very strong attack
scenario. `Ori' means the original model. `Rand' means adding some randomization layers. `Tri. Reg' means applying our triplet regularization.}
\end{table*}

\subsection{Defense-GAN}

Defense-GAN used Vanishing \& Exploding Gradients. This defense do not affect the training and testing of classifier. The adversarial training process was even used in the original paper and got a well performance. So we apply our triplet regularization under the same setting of parameters and the result shows that the new loss can improve the robustness based on the original defense.

The parameters in the triplet loss function are set to be $\lambda_1 = 0.3, \lambda_2=0.5, \alpha = 2.0$, and Defense-GAN has $L = 200$ and $R = 10$. The results are shown in Table 14 and the model structures are listed in Table 15. 
\begin{table*}[!htbp]
\centering
\scalebox{0.9}[0.9]{
\begin{tabular}{|c|c|c|c|c|c|c|c|}
\hline
Attack                & Model & \begin{tabular}[c]{@{}l@{}}No\\ Attack\end{tabular} & \begin{tabular}[c]{@{}l@{}}No\\ Defense\end{tabular} & \begin{tabular}[c]{@{}l@{}}Defense-\\ GAN-Rec\end{tabular} & \begin{tabular}[c]{@{}l@{}}Adv.\\ Train\end{tabular} & \begin{tabular}[c]{@{}l@{}}AT$^2$L\end{tabular} & \begin{tabular}[c]{@{}l@{}}Defense-\\ GAN-Rec\\ + Tri. Reg\end{tabular} \\ \hline
\multirow{4}{*}{FGSM} & A     & 0.3                                                 & 88.3                                                 & 1.2                                                        & 34.9                                                    & 23.6                                                                & \textbf{1.1}                                                                             \\
                      & B     & 3.8                                                 & 97.8                                                 & 4.4                                                        & 94.0                                                    & 96.8                                                                & \textbf{0.7}                                                                             \\
                      & C     & 0.4                                                 & 67.9                                                 & 1.1                                                        & 21.4                                                    & 11.9                                                                & \textbf{0.8}                                                                             \\
                      & D     & 0.8                                                 & 96.2                                                 & 2.0                                                        & 26.8                                                    & 15.2                                                                & \textbf{1.6}                                                                             \\ \hline
\multirow{4}{*}{C\&W}   & A     & 0.3                                                 & 85.9                                                 & \textbf{1.1}                                                        & 92.3                                                    & 96.5                                                                & 1.4                                                                             \\
                      & B     & 3.8                                                 & 96.8                                                 & 8.4                                                        & 72.0                                                    & 70.7                                                                & \textbf{4.7}                                                                             \\
                      & C     & 0.4                                                 & 87.4                                                 & \textbf{1.1}                                                        & 96.9                                                    & 93.7                                                                & \textbf{1.1}                                                                             \\
                      & D     & 0.8                                                 & 96.8                                                 & 1.7                                                        & 99.0                                                    & 87.7                                                                & \textbf{1.4}                                                                             \\ \hline
\end{tabular}}
\caption{Classification error rates of different classifier models using various defense strategies on
the MNIST datasets, under FGSM and C\&W known type of
attacks. `Adv. Train' mean a traditional adversarial training process. `Tri. Reg' means applying our triplet regularization.}
\end{table*}

\begin{table*}[!htbp]
\centering
\scalebox{0.85}[0.9]{
\begin{tabular}{llllll}
A                                   & B                                    & C                                    & D                                   & Generator                             & Discriminator   \\ \hline
\multicolumn{1}{c|}{Conv(64,5x5,1)} & \multicolumn{1}{c|}{Dropout(0.2)}    & \multicolumn{1}{c|}{Conv(128,3x3,1)} & \multicolumn{1}{c|}{FC(200)}        & \multicolumn{1}{c|}{FC(4096)}         & Conv(64,5x5,2)  \\
\multicolumn{1}{c|}{ReLU}           & \multicolumn{1}{c|}{Conv(64,8x8,2)}  & \multicolumn{1}{c|}{ReLU}            & \multicolumn{1}{c|}{ReLU}           & \multicolumn{1}{c|}{ReLU}             & LeakyReLU(0.2)  \\
\multicolumn{1}{c|}{Conv(64,5x5,2)} & \multicolumn{1}{c|}{ReLU}            & \multicolumn{1}{c|}{Conv(64,3x3,2)}  & \multicolumn{1}{c|}{Dropout(0.5)}   & \multicolumn{1}{c|}{ConvT(256,5x5,1)} & Conv(128,5x5,2) \\
\multicolumn{1}{c|}{ReLU}           & \multicolumn{1}{c|}{Conv(128,6x6,2)} & \multicolumn{1}{c|}{ReLU}            & \multicolumn{1}{c|}{FC(200)}        & \multicolumn{1}{c|}{ReLU}             & LeakyReLU(0.2)  \\
\multicolumn{1}{c|}{Dropout(0.25)}  & \multicolumn{1}{c|}{ReLU}            & \multicolumn{1}{c|}{Dropout(0.25)}   & \multicolumn{1}{c|}{ReLU}           & \multicolumn{1}{c|}{ConvT(128,5x5,1)} & Conv(256,5x5,2) \\
\multicolumn{1}{c|}{FC(128)}        & \multicolumn{1}{c|}{Conv(128,5x5,1)} & \multicolumn{1}{c|}{FC(128)}         & \multicolumn{1}{c|}{Dropout(0.5)}   & \multicolumn{1}{c|}{ReLU}             & LeakyReLU(0.2)  \\
\multicolumn{1}{c|}{ReLU}           & \multicolumn{1}{c|}{ReLU}            & \multicolumn{1}{c|}{ReLU}            & \multicolumn{1}{c|}{FC(10)+Softmax} & \multicolumn{1}{c|}{ConvT(1,5x5,1)}   & FC(1)           \\
\multicolumn{1}{c|}{Dropout(0.5)}   & \multicolumn{1}{c|}{Dropout(0.5)}    & \multicolumn{1}{c|}{Dropout(0.5)}    & \multicolumn{1}{c|}{}               & \multicolumn{1}{c|}{Sigmoid}          & Sigmoid         \\
\multicolumn{1}{c|}{FC(10)+Softmax} & \multicolumn{1}{c|}{FC(10)+Softmax}  & \multicolumn{1}{c|}{FC(10)+Softmax}  & \multicolumn{1}{c|}{}               & \multicolumn{1}{c|}{}                 &                
\end{tabular}}
\caption{Neural network architectures used for classifiers, substitute models and GANs.
}
\end{table*}

\end{document}